%% file: main.tex
\documentclass[10pt,journal,compsoc]{IEEEtran}

\ifCLASSOPTIONcompsoc
  \usepackage[nocompress]{cite}
\else
  \usepackage{cite}
\fi
\usepackage[hyphens]{url}  
\usepackage{graphicx} 
\usepackage{algpseudocode,algorithm,algorithmicx}

\usepackage{caption}

\usepackage{booktabs} 
\usepackage{amsmath}
\usepackage{amssymb}
\usepackage{multirow}
\usepackage{url}
\usepackage{enumerate}
\usepackage{bm}
\usepackage{subcaption}
\usepackage{CJKutf8}
\usepackage{xspace}
\usepackage{xcolor}

\newcommand{\eg}{\textit{e}.\textit{g}.\@\xspace}
\newcommand{\ie}{\textit{i}.\textit{e}.\@\xspace}
\newcommand{\etal}{\textit{et al}.\@\xspace}

\newcommand{\eua}{\textit{ES Attack}\xspace}

\DeclareMathOperator*{\argmin}{arg\,min}

\newcommand{\xy}[1]{{\color{black} #1}\normalfont}
\newcommand{\xyy}[1]{{\color{black} #1}}

\begin{document}

\title{ES Attack: Model Stealing against Deep Neural Networks without Data Hurdles\thanks{© 2022 IEEE.  Personal use of this material is permitted.  Permission from IEEE must be obtained for all other uses, in any current or future media, including reprinting/republishing this material for advertising or promotional purposes, creating new collective works, for resale or redistribution to servers or lists, or reuse of any copyrighted component of this work in other works.}}

\author{\IEEEauthorblockN{Xiaoyong Yuan\IEEEauthorrefmark{1}, Leah Ding\IEEEauthorrefmark{2}, Lan Zhang\IEEEauthorrefmark{1}, Xiaolin Li\IEEEauthorrefmark{3}, and Dapeng Wu\IEEEauthorrefmark{4},~\IEEEmembership{Fellow,~IEEE}}\\
\IEEEauthorblockA{
\IEEEauthorrefmark{1}Michigan Technological University
\IEEEauthorrefmark{2}American University
\IEEEauthorrefmark{3}Cognization Lab
\IEEEauthorrefmark{4}University of Florida
}
}

\IEEEtitleabstractindextext{%
\begin{abstract}
Deep neural networks (DNNs) have become the essential components for various commercialized machine learning services, such as Machine Learning as a Service (MLaaS). 
Recent studies show that machine learning services face severe privacy threats -  well-trained DNNs owned by MLaaS providers can be stolen through public APIs, namely model stealing attacks. 
However, most existing works undervalued the impact of such attacks, where a successful attack has to acquire confidential training data or auxiliary data regarding the victim DNN.
In this paper, we propose \eua, a novel model stealing attack without any data hurdles. 
By using heuristically generated synthetic data, \eua iteratively trains a substitute model and eventually achieves a functionally equivalent copy of the victim DNN. 
The experimental results reveal the severity of \eua: i) \eua successfully steals the victim model without data hurdles, and \eua even outperforms most existing model stealing attacks using auxiliary data in terms of model accuracy; ii) most countermeasures are ineffective in defending \eua; iii) \eua facilitates further attacks relying on the stolen model.
\end{abstract}

\begin{IEEEkeywords}
model stealing, deep neural network, knowledge distillation, data synthesis.
\end{IEEEkeywords}
}

\maketitle
\IEEEdisplaynontitleabstractindextext
\IEEEpeerreviewmaketitle

\IEEEraisesectionheading{\section{Introduction}}
\input{1-intro.tex}

\section{Problem Statement}
\input{2-background.tex}

\section{\eua}
\input{3-steal.tex}

\section{Evaluation of \eua}
\input{4-eval.tex}

\subsection{Further Attacks: A Case Study on Black-Box Adversarial Attacks}
\label{sec:adv}
\input{adversarial.tex}

\section{Countermeasures of Model Stealing}
\label{sec:defense}
\input{5-defenses.tex}

\section{Related Work}
\label{sec:related}
\input{6-related.tex}

\section{Conclusion}
\input{8-conclusion.tex}

\section*{Acknowledgments}
This work was supported in part by National Science Foundation (CCF-2007210, CCF-2106610, and CCF-2106754).
\bibliographystyle{IEEEtran}
\bibliography{deep,intro,steal}

\clearpage
\onecolumn

\section*{Appendix}
\input{9-appendix.tex}

\end{document}

%% file: 1-intro.tex
As one of the typical business models, Machine-Learning-as-a-Service (MLaaS) provides a platform to facilitate users to use machine learning models~\cite{hunt2018chiron}. Users can access well-trained machine learning models via public APIs provided by MLaaS providers, without building a model from scratch. MLaaS allows users to query machine learning models in the form of pay-per-query and get responses of the model's predictions.
Recent studies show that machine learning services face severe privacy threats:
model stealing attacks steal functionally equivalent copies from MLaaS providers through multiple queries~\cite{tramer2016stealing,orekondy2019knockoff,correia2018copycat,pal2019framework,yu2020cloudleak}. 
Model stealing attacks exploit the tensions between queries and their corresponding feedback, \ie, the output predictions. 
Tramer~\etal extract machine learning model's parameters by solving equations derived from the model architecture~\cite{tramer2016stealing}. 
However, it requires the exact knowledge of ML architectures and becomes difficult to scale up to steal deep neural networks (DNNs)~\cite{papernot2017practical}. 
Existing model stealing attacks against DNNs require specific knowledge of the model’s training data: the exact training data \cite{wang2018stealing,yu2020cloudleak}, seed samples from the training data\cite{juuti2018prada}, and auxiliary data that shares similar attributes as the training data or within the same task domain~\cite{orekondy2019knockoff,correia2018copycat,pal2019framework}. 

Most existing model stealing attacks require the knowledge of training data or auxiliary data regarding the victim deep neural networks, which undervalued the impact of model stealing attacks. 
In practice, these data are not always accessible.
Due to recent regulations on data protection (\eg, GDPR and CCPA), many types of personal data are hard to acquire, such as health data and biometric data.
In many domains, companies collect data for their own business and are reluctant to share their data. Government and other organizations usually lack resources and financial supports to create open datasets.
Often, the quality of public data is out-of-date and questionable without updates and maintenance.
The availability of appropriate data protects the victim models from being stolen by the existing model stealing attacks.

\begin{figure}[!tb]
\centering
\includegraphics[width=0.95\linewidth]{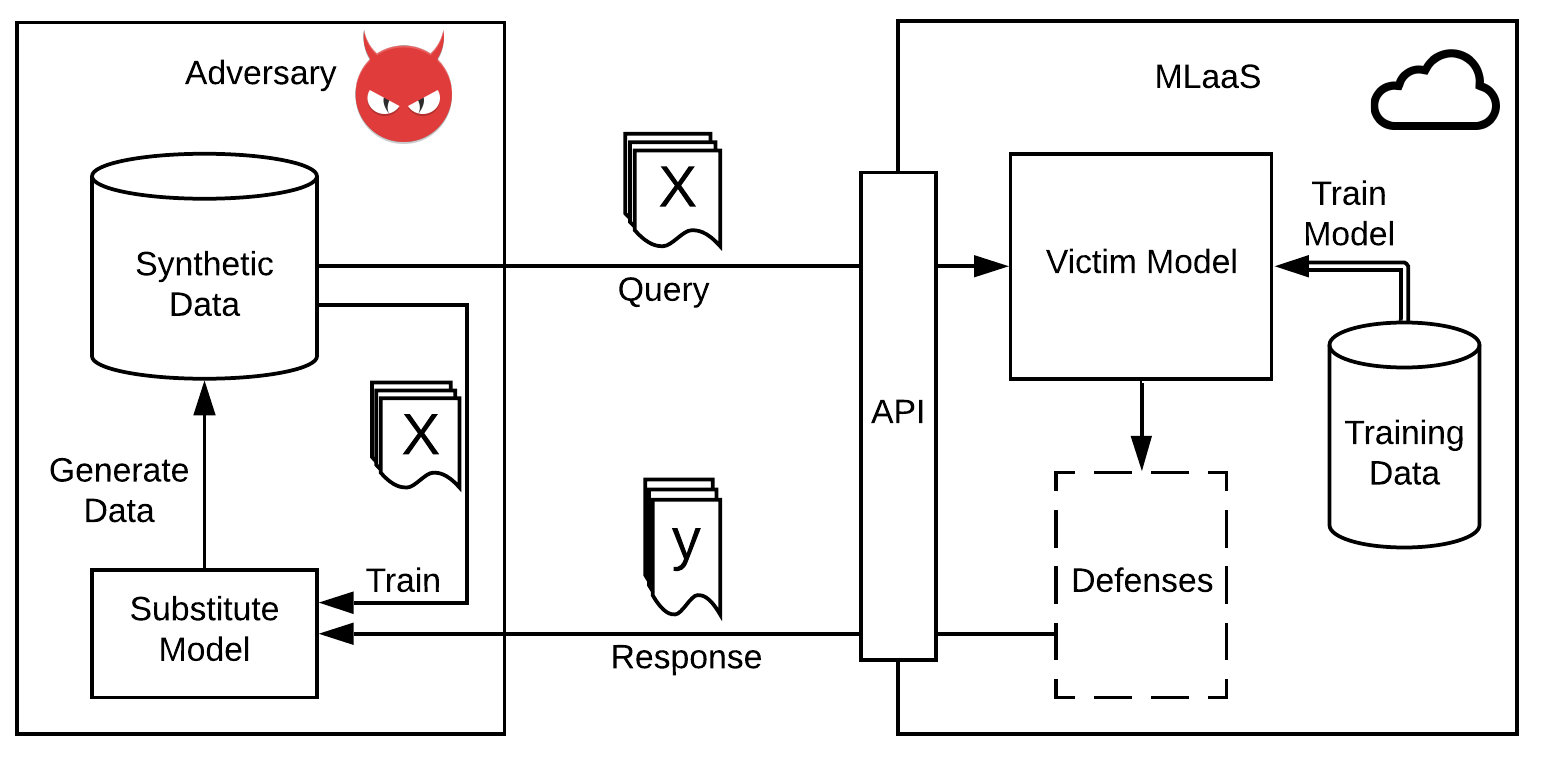}
\vspace{-0.5em}
\caption{\textbf{Diagram of \eua}.}
\label{fig:diagram}
\vspace{-0.5em}
\end{figure}

In this paper, we introduce \eua, a new class of model stealing attacks against DNNs without data hurdles.
\eua heuristically generates synthetic data to overcome the limitations of existing approaches.  
Figure~\ref{fig:diagram} illustrates the diagram of \eua. 
The adversary queries the victim model with the synthetic data $\bm{x}$ and labels the data using responses $\bm{y}$ via the MLaaS provider's APIs. 
The MLaaS provider may deploy defenses to prevent model leakage.
A substitute model is iteratively trained using the synthetic data of $\bm{x}$ with corresponding labels $\bm{y}$ and eventually approximates the functionality of the victim model.
There are two key steps in \eua: E-Step to Estimate the parameters in a substitute model and S-Step to Synthesize the dataset for attacking.

Compared to existing model stealing attacks, \eua does not require i) information about the internals of the victim's model (\eg, its architecture, hyper-parameters, and parameters), and ii) prior knowledge of the victim model's training data.
The adversary only observes responses given by the victim model. 


In summary, our contributions are fourfold:
\begin{enumerate}[1)]
    \item We propose a novel model stealing framework \eua that does not require any knowledge of the victim model's training data. Compared to model stealing using auxiliary datasets, our proposed \eua improves the model accuracy by 44.57\%.
    \item \eua generates better synthetic datasets compared with the auxiliary datasets in terms of quality and diversity.
    \item  We demonstrate that the stolen model successfully facilitates black-box adversarial attacks against the victim model. 
    \item Three investigated countermeasures are not effective in preventing \eua.
\end{enumerate}

%% file: 2-background.tex
\xy{In this paper, we consider a typical service in Machine-Learning-as-a-Service (MLaaS), \ie, image classification. 
Given a task domain $\mathcal{T}$, an MLaaS provider (\ie, victim) collects a training dataset $\mathcal{D}_{\text{\text{train}}}$ and a test dataset $\mathcal{D}_{\text{test}}$, consisting of a set of images and their corresponding labels $\{(\bm{x}, \bm{l})\}$. 
Further, the MLaaS provider trains a machine learning model (\ie, victim model) $f_v$ on private training dataset $\mathcal{D}_{\text{train}}$ and provides the image classification services to the public using the victim model.
Normal users can access the trained model $f_v$ by querying data sample $\bm{x}$ and get the response from the victim model regarding the predicted probabilities of $K$ classes $\bm{y} = f_v(\bm{x})$ . 

The goal of model stealing adversaries is to build a model (\ie, stolen model) $f_s$ that is functionally equivalent to the victim model $f_v$. We assume that the adversary can query the victim model by pretending themselves as normal users.
Most model stealing attacks assume that the adversary has full or partial prior knowledge of the victim model's training data $\mathcal{D}_{\text{train}}$. 

\xyy{In this paper, we consider a more realistic scenario that the adversary cannot access the victim's private training data.
Specifically, we assume that the adversary does not know the victim's training data $\mathcal{D}_{\text{train}}$ or any auxiliary data related to $\mathcal{D}_{\text{train}}$.}

To evaluate the risk of model stealing attacks, we leverage the prediction accuracy on the test dataset $\mathcal{D}_{\text{test}}$. 
By mimicking the behavior of the victim model, the adversary aims to achieve good performance on the unknown test dataset $\mathcal{D}_{\text{test}}$ using the stolen model $f_s$. 
}

%% file: 3-steal.tex
In this section, we present the design of \eua, and propose two heuristic approaches for data synthesis.

\subsection{Design of ES Attack}

\begin{figure*}[!tb]
\centering
\begin{subfigure}{0.28\textwidth}
\centering
\includegraphics[width=\linewidth]{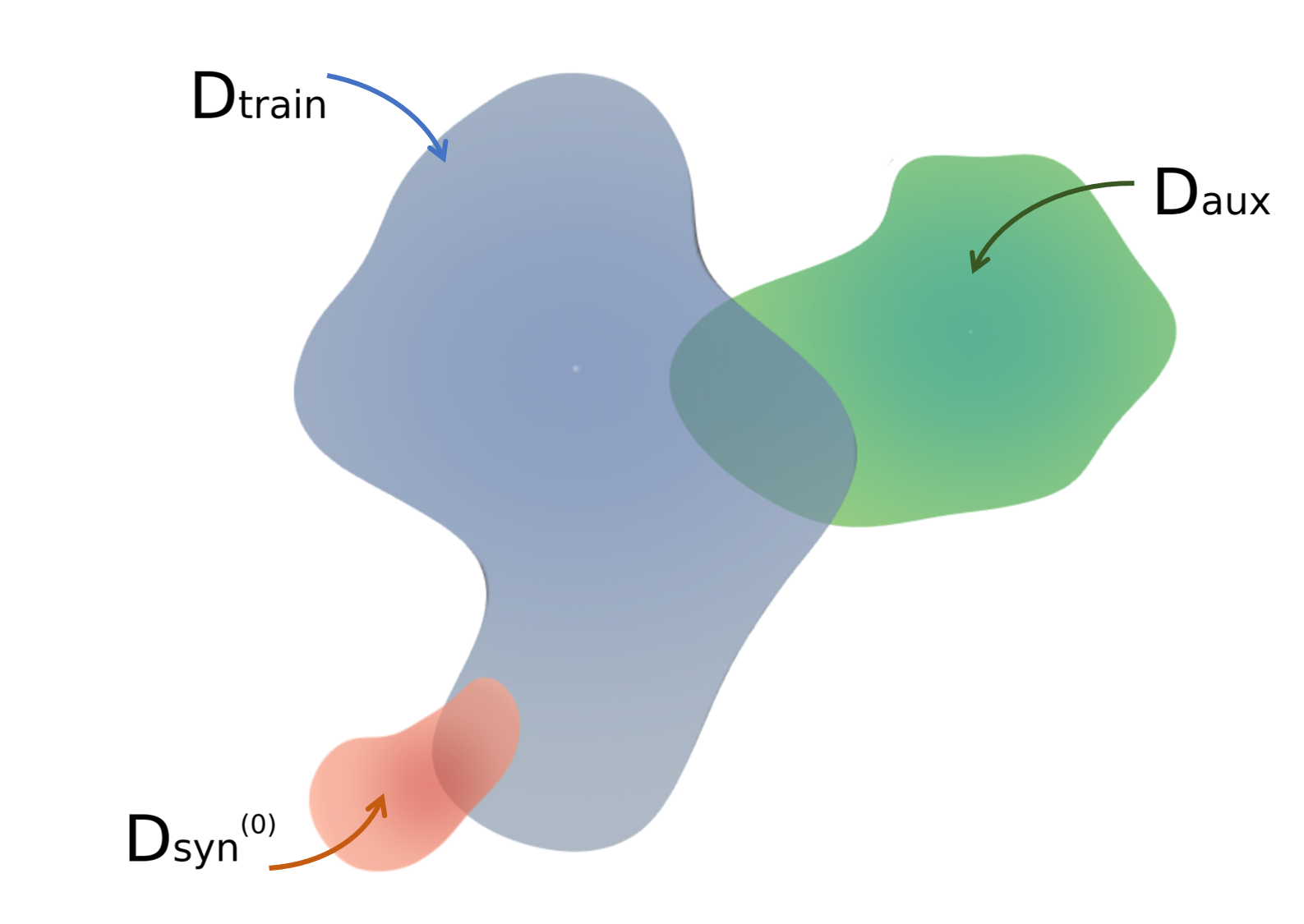}
\caption{Initial Steal ($t=0$)}
\end{subfigure}
\begin{subfigure}{0.28\textwidth}
\centering
\includegraphics[width=\linewidth]{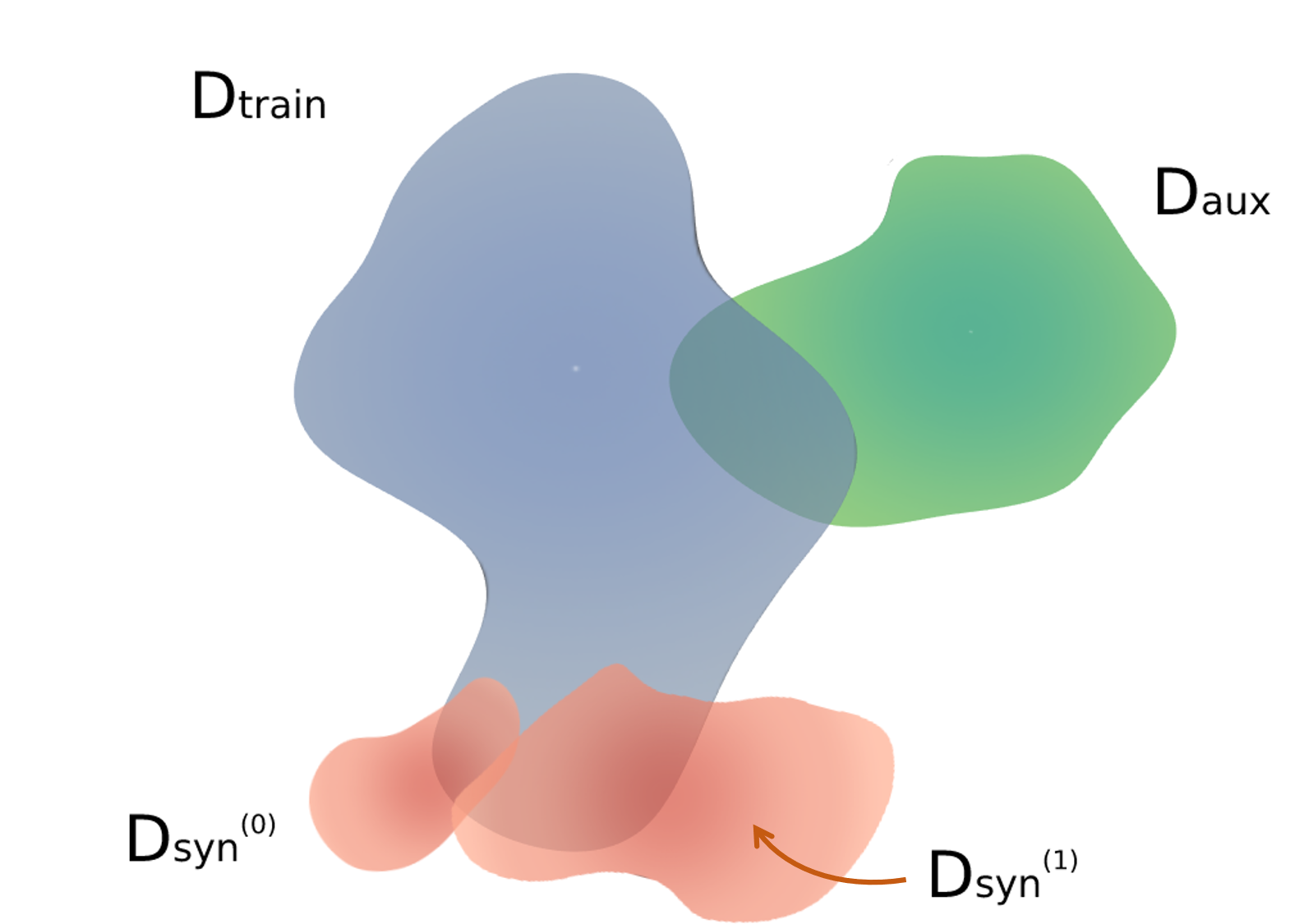}
\caption{The $1$st Steal ($t=1$)}
\end{subfigure}
\begin{subfigure}{0.28\textwidth}
\centering
\includegraphics[width=\linewidth]{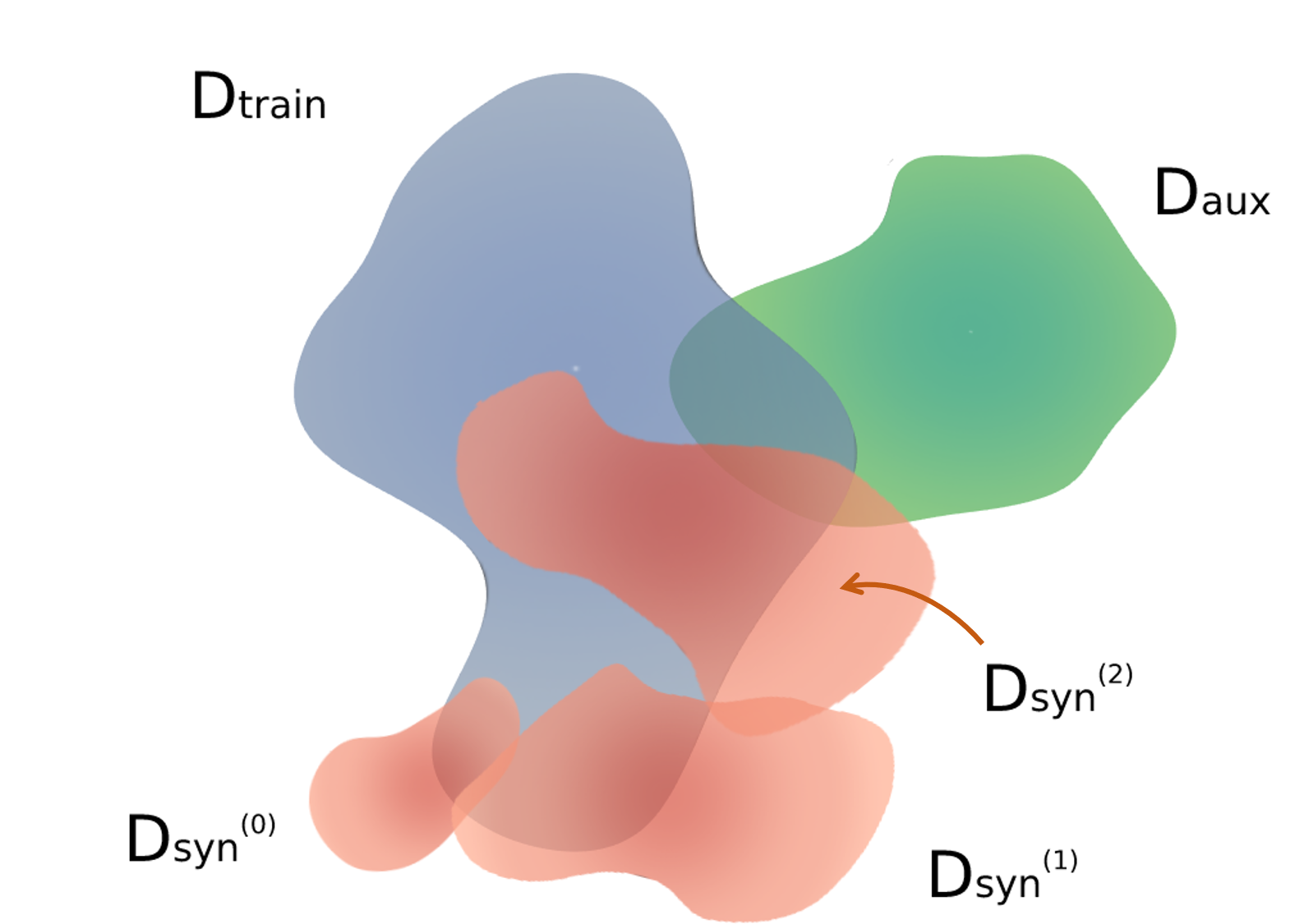}
\caption{The $2$st Steal ($t=2$)}
\end{subfigure}
\begin{subfigure}{0.28\textwidth}
\centering
\includegraphics[width=\linewidth]{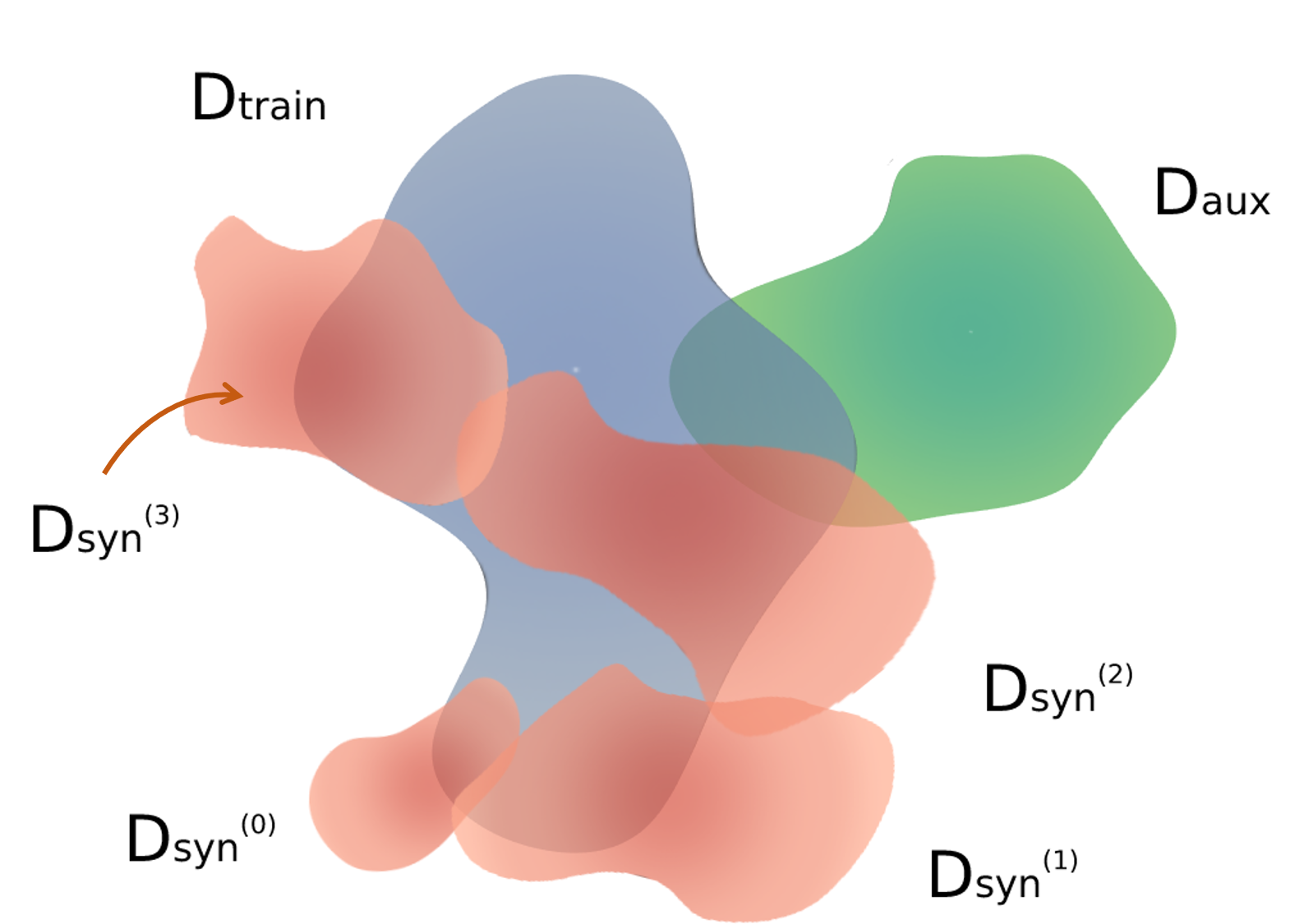}
\caption{The $3$rd Steal ($t=3$)}
\end{subfigure}
\begin{subfigure}{0.28\textwidth}
\centering
\includegraphics[width=\linewidth]{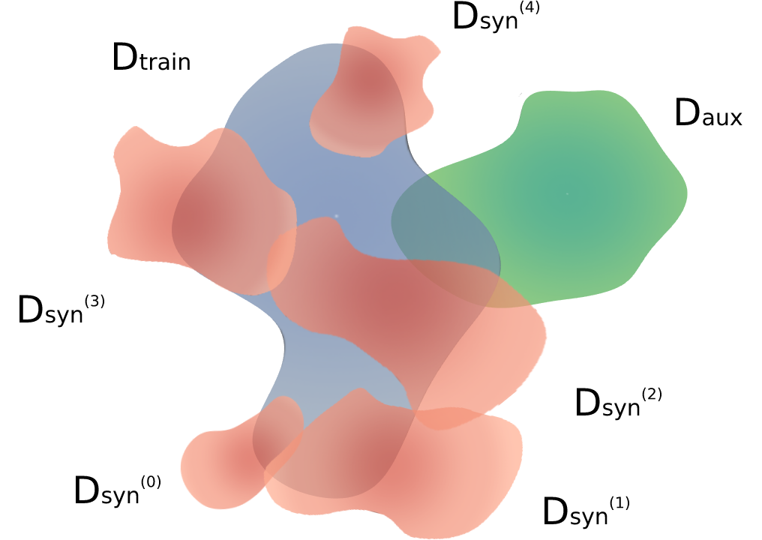}
\caption{The $4$th Steal ($t=4$)}
\end{subfigure}
\begin{subfigure}{0.28\textwidth}
\centering
\includegraphics[width=\linewidth]{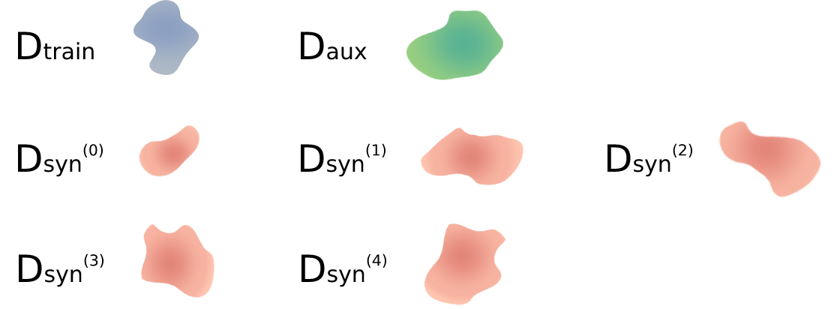}
\end{subfigure}
\vspace{-0.5em}
\caption{\textbf{Progress of Data Synthesis During \eua.} We compare the synthetic datasets $\mathcal{D}_{\text{syn}}^{(t)}$ (in red), generated by our proposed attack, with the victim's training dataset $\mathcal{D}_{\text{train}}$ (in blue) and the auxiliary dataset $\mathcal{D}_{\text{aux}}$ (in green). The auxiliary dataset $\mathcal{D}_{\text{aux}}$ may share similar input space with the victim's training dataset $\mathcal{D}_{\text{train}}$, but a large space of $\mathcal{D}_{\text{train}}$ cannot be covered by the auxiliary dataset. 
In our attack, we first initial a randomly generated synthetic dataset $\mathcal{D}_{\text{syn}}^{(0)}$. $\mathcal{D}_{\text{syn}}^{(0)}$ share some attributes with $\mathcal{D}_{\text{train}}$, which might be less than $\mathcal{D}_{\text{aux}}$. During our attacks, $\mathcal{D}_{\text{syn}}^{(t)} (t=1, 2, 3)$ get information from our substitute models and explore more space in $\mathcal{D}_{\text{train}}$ in each iteration. The goal of the attacks is to cover the input space of $\mathcal{D}_{\text{train}}$ as much as possible. 
Note that the adversary trains the substitute model on a synthetic dataset $\mathcal{D}_{\text{syn}}^{(i)}$, ($i=0, 1, \cdots, N$) in each stealing epoch. As a sum, the substitute model is trained on all the synthetic datasets.
}
\label{fig:steal}
\vspace{-0.5em}
\end{figure*}

Model stealing attacks aim to build a model $f_s$ that is functionally equivalent to the victim model $f_v$.
Theoretically, if the adversary can train the substitute model on all the samples in the input space of $f_v$, the substitute model can achieve the same performance as the victim model. However, it is infeasible to query all the samples in the input space. 
Therefore, a critical step for model stealing attacks is to explore the input sub-space that represents the task domain $\mathcal{T}$.  
Adversaries will mimic the behavior of victim models in the input sub-space. \cite{orekondy2019knockoff,correia2018copycat,pal2019framework} leverage public datasets as an auxiliary dataset to train the substitute model to approximate the output of the victim model. The auxiliary data share common attributes with $\mathcal{D}_{\text{train}}$, which can be used to train the substitute model. 
However, these approaches are not practical due to many reasons: 
i) Data with shared attributes is not always available.
Confidential data such as medical images, financial records are not publicly available. 
The scarcity of data is still a critical problem in many domains. 
ii) The relationship between the available auxiliary data on the public domain and the task domain $\mathcal{T}$ is unclear, which brings a challenge to select a proper auxiliary dataset. The rationale for selecting a specific auxiliary dataset is missing in most of the existing approaches. 
In the experiments, we show that using a randomly generated dataset, a special case of an auxiliary dataset, fails to steal the victim model.
iii) The quality of data used for training the substitute model cannot be improved during model stealing. 
The data samples are fixed in the auxiliary dataset. 

Therefore, we propose an \eua to heuristically explore the potential input space related to task domain $\mathcal{T}$ by learning from the victim model. 
We outline \eua in Algorithm~\ref{alg:attack}.
First, \eua initializes a randomly synthetic dataset $\mathcal{D}_{\text{syn}}^{(0)}$, which may share few attributes with $\mathcal{D}_{\text{train}}$, most likely fewer than $\mathcal{D}_{\text{aux}}$. Second, \eua trains a substitute model $f_s$ based on the samples from the synthetic dataset and their predictions from the victim model. 
Then, \eua can generate a better synthetic dataset using the improved substitute model; in the meanwhile, the better synthetic dataset can help improve the substitute model. 
In this way, \eua iteratively synthesizes the datasets and trains the substitute model to improve the quality of the synthetic dataset and the performance of the substitute model. Eventually, the synthetic datasets will approximate the private training dataset, and the substitute model will approximate the victim model or steal the victim model.

Figure~\ref{fig:steal} illustrates the progress of data synthesis during \eua. 
In Figure~\ref{fig:steal}, we compare the synthetic datasets $\mathcal{D}_{\text{syn}}^{(t)}$ (in red) with the victim's training dataset $\mathcal{D}_{\text{train}}$ (in blue) and the auxiliary dataset $\mathcal{D}_{\text{aux}}$ (in green). $\mathcal{D}_{\text{aux}}$ may share similar input space with $\mathcal{D}_{\text{train}}$, but in most cases, adversaries may not know the distance between the distribution of $\mathcal{D}_{\text{aux}}$ and the distribution of $\mathcal{D}_{\text{train}}$. 
Hence, $\mathcal{D}_{\text{train}}$ may not be fully covered by $\mathcal{D}_{\text{aux}}$.
However, after initializing the synthetic dataset $\mathcal{D}_{\text{syn}}^{(0)}$, \eua will iteratively improve the synthetic datasets $\mathcal{D}_{\text{syn}}^{(t)} (t=1, 2, 3, 4, \ldots)$, and explore more space in the training dataset $\mathcal{D}_{\text{train}}$. 

Two key steps in \eua, E-Step and S-Step, are described as follows.
\par\noindent
\textbf{E-Step: } 
Estimate parameters in the substitute model on the synthetic dataset using knowledge distillation. The knowledge distillation approach transfers the knowledge from $f_v$ to $f_s$ with minimal performance degradation:
\begin{equation}
\label{eq:estep}
f_s^{(t)} \gets \argmin_{f_s} \mathrm{L}_{\text{KD}}(f_s, f_v;\mathcal{D}_{\text{syn}}^{(t-1)}), 
\end{equation}
where $f_s^{(t)}$ denotes the substitute model at iteration $t$ and $\mathcal{D}_{\text{syn}}^{(t-1)}$ denotes the synthetic dataset at the previous iteration $t-1$.
The objective function $\mathrm{L}_{\text{KD}}$ is defined as knowledge distillation loss to make $f_s^{(t)}$ approximate the victim model $f_v$:
\begin{equation}
    \mathrm{L}_{\text{KD}}(f_s, f_v; \mathcal{D}_{\text{syn}}) = \frac{1}{|\mathcal{D}_{\text{syn}}|}\sum_{\bm{x} \in \mathcal{D}_{\text{syn}}}\mathrm{L}_{\text{CE}}(f_s(\bm{x}), f_v(\bm{x})),
\end{equation}
where $L_{\text{\text{CE}}}$ denotes the cross-entropy loss. We train the substitute model by minimizing the objective function (Equation~\ref{eq:estep}) for $M$ epochs using Adam~\cite{kingma2014adam}. 

\par\noindent
\textbf{S-Step: } Synthesize the dataset $\mathcal{D}_{\text{syn}}^{(t)} = \{\bm{x}\}$ consisted of the synthetic input samples.


\begin{algorithm}[!tb]
\caption{\eua}
\label{alg:attack}
\textbf{INPUT}: \\
\text{\quad The black-box victim model $f_v$}\\
\text{\quad Number of classes $K$}\\
\text{\quad Number of stealing epochs $N$}\\
\text{\quad Number of training epochs for each stealing epoch $M$}\\
\textbf{OUTPUT}: \\
\text{\quad The substitute model $f_s^{(N)}$}
\begin{algorithmic}[1]
\State Initialize a synthetic dataset $\mathcal{D}_{\text{syn}}^{(0)}$ by randomly sampling $\bm{x}$ from a Gaussian distribution.
\State Construct an initial substitute model $f_s^{(0)}$ by initializing the parameters in the model.
\For{$t \gets 1$ to $N$}
    \State \textbf{E-Step:} \xy{Estimate the parameters in the substitute model $f_s^{(t)}$ using knowledge distillation for $M$ epochs on the synthetic dataset $\mathcal{D}_{\text{syn}}^{(t-1)}$}. 
    \State \textbf{S-Step:} Synthesize a new dataset $\mathcal{D}_{\text{syn}}^{(t)}$ based on the knowledge of the substitute model $f_s^{(t)}$.
\EndFor
\State \textbf{return} $f_s^{(N)}$.
\end{algorithmic}
\end{algorithm}


\subsection{Two approaches for Data Synthesis (S-Step)}
\label{sec:data_syn}
\xy{
Data synthesis (S-step in \eua) aims to explore the possible data that reveal the data distribution of the victim's training dataset and benefit the substitute model. 
The most challenging problem in data synthesis is the lack of gradient of the victim model. The adversary can only get the prediction rather than the gradient from the victim model. Accordingly, the data generated by the adversary cannot be tuned by the victim model directly. The existing approaches used in data-free knowledge distillation fail to be applied in the model stealing attacks, since they require to back-propagate the gradients from the victim model. More discussion about the difference between model stealing and data-free knowledge distillation can be found in Section~\ref{sec:related_knowledge}.
To overcome this challenge, we propose two data synthesis approaches that make use of the gradients of the substitute model as a proxy for updating the synthetic input data. 

Specifically, we introduce two approaches to generate the synthetic data: namely, \textit{DNN-SYN} and \textit{OPT-SYN}. 
Both approaches start with generating a set of random labels and initial data samples. Further the approaches update the data samples based on the assigned labels and the gradients from the substitute model. Once the substitute model is close to the victim model, the synthetic data becomes close to the distribution of the victim's training dataset.
}

\subsubsection{DNN-SYN.}
\begin{figure}[!t]
\centering
\includegraphics[width=0.35\linewidth]{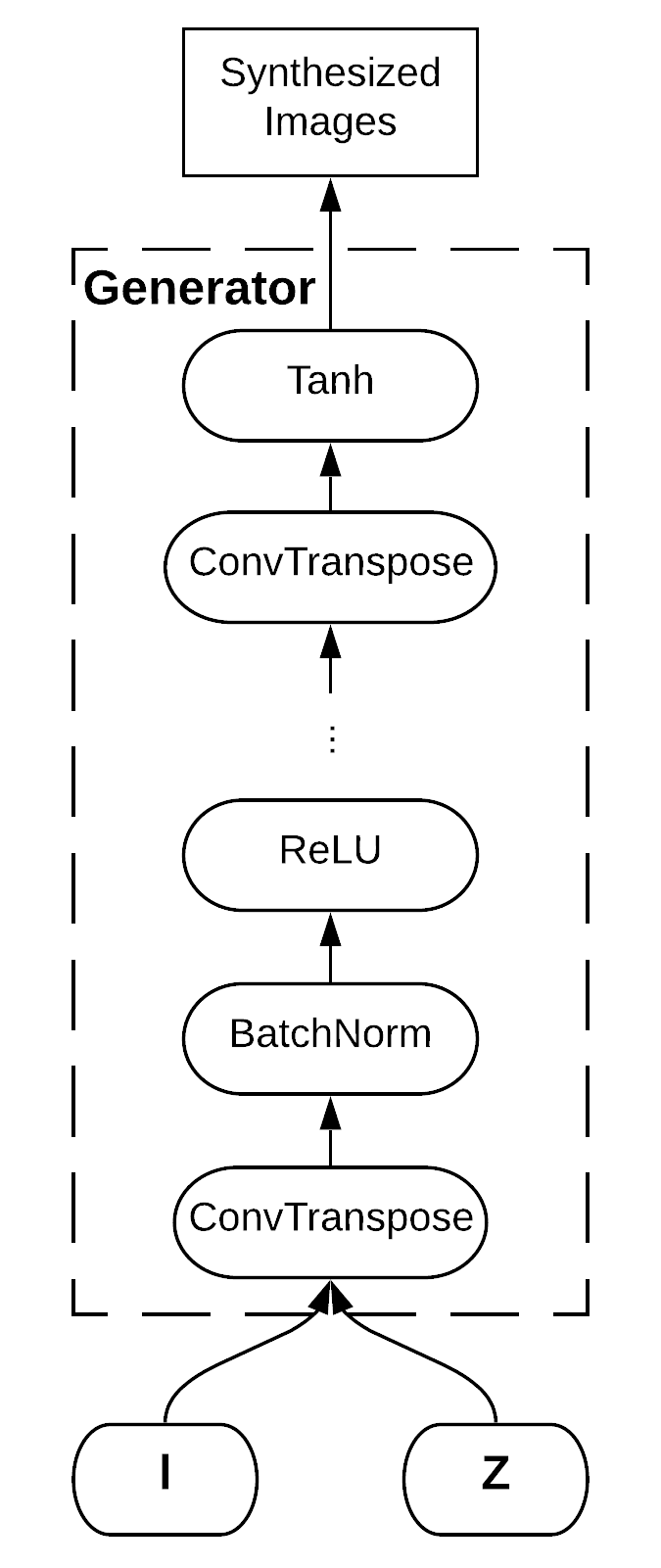}
\vspace{-0.5em}
\caption{\textbf{DNN generator $G$ in \textit{DNN-SYN}.}}
\label{fig:acgan}
\vspace{-1em}
\end{figure}

We design a DNN-based generator to synthesize images that can be classified by our substitute model with high confidence. 
The design of image generator $G$ follows the major architecture of Auxiliary Classifier GAN (ACGAN)~\cite{odena2017conditional}, a variant of Generative Adversarial Network (GAN), which can generate images with label conditioning. 
We refer to the data generation approach using DNN as \textit{DNN-SYN}.

We describe the procedure of \textit{DNN-SYN} as follows: 
\begin{enumerate}[\textbf{Step} 1:]
\item Randomly assign a set of labels $\bm{l}=\{\bm{l_1}, \bm{l_2}, \ldots, \bm{l_n}\}$, where $\bm{l_i}$ denotes a $K$-dimensional one-hot vector. 

\item Train a DNN generator $G$ with parameter $w_G$ to generate data from a random latent vector $\bm{z_i}$. $G$ is optimized that the generated data can be classified by $f_s$ as assigned labels $L$ with high confidence:
\begin{equation}
\label{eq:gan}
\min_{w_G} \quad \mathrm{L_{\text{img}}}(G, \bm{l}) \buildrel \text{def}\over = \sum_{i}^n \mathrm{L}_{\text{CE}}(f_s(G(\bm{z_i}, \bm{l_i})), \bm{l_i}).
\end{equation}

\item Generate a synthetic dataset using the generator trained in Step 2: $\mathcal{D}_{\text{syn}} = \{G(\bm{z_i}, \bm{l_i})\}$.
\end{enumerate}

In addition, mode collapse is one of the critical problems for training GANs. To avoid mode collapse in \textit{DNN-SYN}, we use a mode seeking approach to increase the diversity of data samples~\cite{mao2019mode}. Mode seeking has been shown simple yet effective in mitigating mode collapse.
We generate two images $G(\bm{z_i^1})$ and $G(\bm{z_i^2})$ using latent vectors $\bm{z_i^1}$ and $\bm{z_i^2}$ and maximize the ratio of the distance of images to the distance of latent vectors. In other words, we minimize the mode seeking loss $\mathrm{L}_{\text{ms}}$:
\xy{
\begin{equation}
    \mathrm{L}_{\text{ms}}(G, \bm{l}) \buildrel \text{def}\over = \sum_{i}^n \frac{d(\bm{z_i^1}, \bm{z_i^2})}{d(G(\bm{z_i^1}, \bm{l_i}), G(\bm{z_i^2}, \bm{l_i}))},
\end{equation}
}
where $d(\cdot, \cdot)$ denotes a distance metric. In this paper, we use $\ell_2$ norm distance.
We sum up the original objective function $\mathrm{L}_{\text{img}}$ and the regularization term  $\mathrm{L}_{\text{ms}}$ and minimize the new objective function:
\begin{equation}
    \mathrm{L_{\text{DNN}}} = \mathrm{L}_{\text{img}} + \lambda \mathrm{L}_{\text{ms}},
\end{equation}
where $\lambda$ denotes the hyper-parameter to adjust the value of regularization. In the experiment, we set $\lambda$ as $1$. 

Figure~\ref{fig:acgan} illustrates the architecture of DNN-based generator in \textit{DNN-SYN}. 
We follow the major design of ACGAN~\cite{odena2017conditional}.
We feed a random latent vector $\bm{z_i}$ and a generated one-hot label $l_i$ into generator $G$. We concatenate these two vectors and up-sample them using several transposed convolution layers. Transposed convolution layers are parameterized by learnable weights. Each transposed convolution layer is followed by a batch normalization layer and a ReLU activation function except the last transposed convolution layer. 
4 transposed convolution layers are used in the model.
In the final layer, we use a Tanh function to output a synthesis image within $(-1, 1)$.

\subsubsection{OPT-SYN.}

\begin{algorithm}[!tb]
\caption{Data Synthesis of \textit{OPT-SYN}}
\label{alg:opt}
\textbf{INPUT}:\\
\text{\quad The substitute model $f_s^{t}$ at iteration $t$}\\
\text{\quad Number of synthetic data $S$}\\
\text{\quad Number of output classes $K$}\\
\text{\quad Number of optimization iterations $m$}\\
\textbf{OUTPUT}: 
\text{\quad A set of optimized data samples $\mathcal{X}$}
\begin{algorithmic}[1]
\State $\mathcal{X} \gets \emptyset$
\For{$i \gets 1, S$}
    \State // Generate a $K$-dimensional random parameter $\alpha$ from a Gaussian distribution
    \State $\bm{\alpha} \sim \mathcal{N}(0, 1)$
    \State // Sample a prediction vector $\bm{y}$ from a Dirichlet distribution
    \State $\bm{y} \sim D(K, \bm{\alpha})$
    \State // Initialize a data sample $\bm{x}$ from Gaussian distribution
    \State $\bm{x} \sim \mathcal{N}(0, 1)$
    \State // Minimize Equation~\ref{eq:opt} for $m$ iterations
    \State $\bm{x}^* \gets \argmin_{\bm{x}} \quad \mathrm{L}_{\text{CE}}(f_s^{(t)}(\bm{x}), \bm{y})$
    \State // Add $\bm{x}^*$ to set $X$
    \State $\mathcal{X} \gets \mathcal{X} \cup \bm{x}^*$
\EndFor
\State \textbf{return}: $\mathcal{X}$.
\end{algorithmic}
\end{algorithm}

Instead of training a generator to synthesize the dataset, we propose an optimization-based data synthesis approach, \textit{OPT-SYN}, which operates on the input space directly and does not suffer the problem of mode collapse.
In addition, \textit{OPT-SYN} explores a more diverse label space compared to the one-hot labels used in \textit{DNN-SYN}.
\textit{OPT-SYN} first explores the possible prediction vectors $\{\bm{y}\}$ in the task domain $\mathcal{T}$ and then minimizes the cross-entropy loss between $\{\bm{y}\}$ and the substitute model's prediction on the synthetic data:
\begin{equation}
\label{eq:opt}
    \min_{\bm{x}} \mathrm{L}_{\text{CE}}(f_s^{(t)}(\bm{x}), \bm{y}),
\end{equation}
where $f_s^{(t)}$ denotes the substitute model at the $t$th stealing epoch. 
In the experiments, we find that \textit{OPT-SYN} performs better than \textit{DNN-SYN} does in most scenarios.

The proposed \textit{OPT-SYN} approach is detailed in Algorithm~\ref{alg:opt}.
First, to explore the possible prediction vectors, we sample each random vector $\bm{y} = \{y_1, y_2, \ldots, y_K\}$  from a $K$-dimensional Dirichlet distribution with parameter $\bm{\alpha}$. 
Dirichlet distribution is commonly used as conjugate prior distribution of categorical distribution.
From the Dirichlet distribution, we can sample prior probabilities $\{y_1, y_2, \ldots, y_K\}$, where $y_i \in (0, 1)$ and $\sum_{i=1}^K y_i = 1$. $\bm{\alpha}$ is referred to as the concentration parameter, which controls the distribution. The probability density function of Dirichlet distribution $Dir(K, \bm{\alpha})$ can be calculated by:
\begin{equation}
    f(y_1, y_2, \ldots, y_K, \bm{\alpha}) = \frac{1}{B(\bm{\alpha})}\prod_{i=1}^K y_i^{\alpha_i -1},
\end{equation}
where $B(\bm{\alpha})$ denotes the gamma function and $\sum y_1, y_2, \ldots, y_K = 1$.
In the experiment, we randomly sample the parameter $\bm{\alpha}$ from a Gaussian distribution: $\bm{\alpha} \sim \mathcal{N}(0, 1)$ to explore the possible Dirichlet distribution.

Given the prediction vector $\bm{y}$, we synthesize 
data $\bm{x}$ by iteratively minimizing the objective function~\ref{eq:opt}.
The goal is to generate a data sample $\bm{x}^*$ that $f_s^{(t)}$ predicts $\bm{x}^*$ close to $\bm{y}$. 
An adaptive gradient-based optimization algorithm, Adam~\cite{kingma2014adam}, is applied to optimize the objective function iteratively.

%% file: 4-eval.tex
In this section, we evaluate our proposed \eua on three different neural networks and four image classification datasets. 
We compare our results with two baseline attacks. 
Moreover, we investigate the data synthesized during the attacks in terms of data quality and diversity.

\subsection{Experiment Setup}
\subsubsection{Victim Models and Datasets}
In our experiments, we evaluate model stealing attacks on four image classification datasets: MNIST, KMNIST, SVHN, CIFAR-10.
The MNIST dataset~\cite{lecun1998gradient} contains 70,000 28-by-28 gray images of 10 digits. We use 60,000 images for training and 10,000 images for testing following the original train/test split in the MNIST dataset. 
Kuzushiji-MNIST (KMNIST)~\cite{clanuwat2018deep} is a similar dataset to MNIST, containing 70,000 28-by-28 grey images of 10 Hiragana characters. We use 60,000 images for training and 10,000 images for testing.
The SVHN~\cite{netzer2011reading} dataset consists of 60,000 32-by-32 RGB images from house numbers (10 classes from 0 to 9) in the Google Street View dataset.
The CIFAR10~\cite{krizhevsky2009learning} dataset contains 60,000 32-by-32 RGB images with 10 classes. 

We train three types of DNN models on four datasets and use them as the victim models in our experiments. 
We train LeNet5~\cite{lecun1998gradient} on the MNIST~\cite{lecun1998gradient} and KMNIST~\cite{clanuwat2018deep} datasets. 
We train ResNet18~\cite{he2016deep} and ResNet34~\cite{he2016deep} on the SVHN~\cite{netzer2011reading} and CIFAR10~\cite{krizhevsky2009learning} datasets.
LeNet5 is trained for 30 epochs using an SGD optimizer with a learning rate of 0.1 on the MNIST and KMNIST datasets. We train ResNet18 and ResNet34 for 200 epochs with an initial learning rate of 0.1 on the SVHN and CIFAR10 datasets. 
We reduce the learning rate by 10 after 80 and 120 epochs.
We select the models with the highest test accuracies as the victim models.


\subsubsection{Settings of \eua}
For \textit{DNN-SYN}, we input a $100$-dimensional random latent vector and a one-hot label vector into DNN-based generator $G$. 
The substitute model $f_s$ and DNN-based generator $G$ is trained by an Adam optimizer with a learning rate of 0.001. 
$f_s$ and $G$ are trained alternatively for 2,000 epochs each on the MNIST, KMNIST, and SVHN dataset ($N=2000, M=1$), and 15,000 epochs each on the CIFAR10 dataset ($N=15000$, $M=1$). 

For \textit{OPT-SYN}, we synthesize data for $30$ iterations ($M=30$) in each stealing epoch using an Adam optimizer with a learning rate of 0.01. We train the adversary model for $10$ epochs on the synthetic dataset ($M=10$). 
We repeat the stealing for $200$ epochs on the MNIST, KMNIST, and SVHN dataset ($N=200$), and 1,500 epochs on the CIFAR10 dataset ($N=1500$).
To speed up the stealing process, we augment the synthetic dataset by random horizontal flip, horizontal shift, and adding Gaussian noise.


\subsubsection{Baseline Model Stealing Attacks}
We compare \eua with two baseline attacks - model stealing using randomly generated data and auxiliary data.
First, if the adversary has no knowledge of the victim's training data, randomly generated data could be the only dataset the adversary can leverage. 
We form a random dataset with the random data sampled from a Gaussian Distribution $\mathcal{N}(0,1)$ and their prediction from the victim model. 
We train our substitute model using the random dataset iteratively.
Second, we consider public data as an auxiliary dataset. We use data samples from other public datasets and query the victim model with them. We construct an auxiliary dataset and train the substitute model on it. 
To make a fair comparison, we make sure that all the model stealing attacks, including two baseline attacks and two \textit{ES Attacks} (\textit{DNN-SYN} and \textit{OPT-SYN}), train their substitute models for the same epochs.

\begin{table}[!tb]
\renewcommand{\arraystretch}{1.3}
\centering
\caption{Performance comparison of model stealing attacks. 
}
\label{tab:function}
\vspace{-0.5em}
\begin{tabular}{@{}llclc@{}}
\toprule
Dataset & Model & \begin{tabular}[c]{@{}c@{}} Victim \\accuracy (\%)\end{tabular} & Attacks & \begin{tabular}[c]{@{}c@{}} Substitute \\accuracy (\%) \end{tabular} \\ \hline
\multirow{4}{*}{SVHN} & \multirow{4}{*}{ResNet18} & \multirow{4}{*}{95.40}     
& Random & 50.71  \\
&& & Auxiliary & 74.84\\
&& & DNN-SYN & 93.95 \\ 
&& & OPT-SYN & \textbf{93.97}\\ \hline
\multirow{4}{*}{SVHN} & \multirow{4}{*}{ResNet34} & \multirow{4}{*}{95.94}     & Random               & 60.95\\ 
&& & Auxiliary & 82.00 \\ 
&& & DNN-SYN & \textbf{93.34} \\  
&& & OPT-SYN & 93.19\\ \hline
\multirow{4}{*}{CIFAR10} & \multirow{4}{*}{ResNet18}& \multirow{4}{*}{91.12}     & Random               & 11.72\\ 
&& & Auxiliary & 48.73 \\
&& & DNN-SYN & 33.44 \\ 
&& & OPT-SYN & \textbf{84.60}\\ \hline
\multirow{4}{*}{CIFAR10} & \multirow{4}{*}{ResNet34}& \multirow{4}{*}{91.93}     & Random               & 14.45 \\
&& & Auxiliary & 38.55\\
&& & DNN-SYN & 12.69 \\ 
&& & OPT-SYN & \textbf{80.79}\\ \hline
\multirow{4}{*}{MNIST} & \multirow{4}{*}{LeNet5}& \multirow{4}{*}{99.10}     & Random               & 72.18 \\ 
&& & Auxiliary & \textbf{98.96} \\  
&& & DNN-SYN & 91.02 \\  
&& & OPT-SYN & 92.03 \\ \hline
\multirow{4}{*}{KMNIST} & \multirow{4}{*}{LeNet5}& \multirow{4}{*}{95.67}     & Random               & 56.39 \\ 
&& & Auxiliary & 59.43 \\ 
&& & DNN-SYN & 90.37 \\ 
&& & OPT-SYN & \textbf{90.37}\\ \bottomrule
\end{tabular}
\vspace{-0.5em}
\end{table}

\subsection{Performance Evaluation}
\label{eval:func}
We evaluate the performance of \textit{ES Attacks} using two data synthesis approaches and compare them with two baseline attacks. We report the accuracy of model stealing attacks in Table~\ref{tab:function}. 
We compare the results with two baseline attacks that use randomly generated data (Random) and auxiliary data (Auxiliary) to steal the victim model. 

From the evaluation, we observe that \textit{OPT-SYN} can successfully steal the victim models over all the datasets and model architectures. 
Our proposed attacks achieve better performance compared with two baseline attacks. \textbf{On average, \textit{OPT-SYN} improves the best accuracy by 44.57\% compared to the best results of two baseline attacks. }
\textit{OPT-SYN} performs as well as \textit{DNN-SYN} on the SVHN, MNIST, and KMNIST datasets. However, \textit{DNN-SYN} cannot achieve a good performance on the CIFAR10 dataset, which is a more complex dataset and the generator $G$ in \textit{DNN-SYN} may still cause the mode collapse problem.
Both our proposed attacks perform worse than the attacks using auxiliary data (KMNIST) on the MNIST dataset, which suggests that the auxiliary data can be used in the model stealing if the auxiliary data well-represent the target task and the data are available to the adversary.

\begin{figure}[!tb]
\centering
\begin{subfigure}{0.45\linewidth}
\centering
\includegraphics[width=\linewidth]{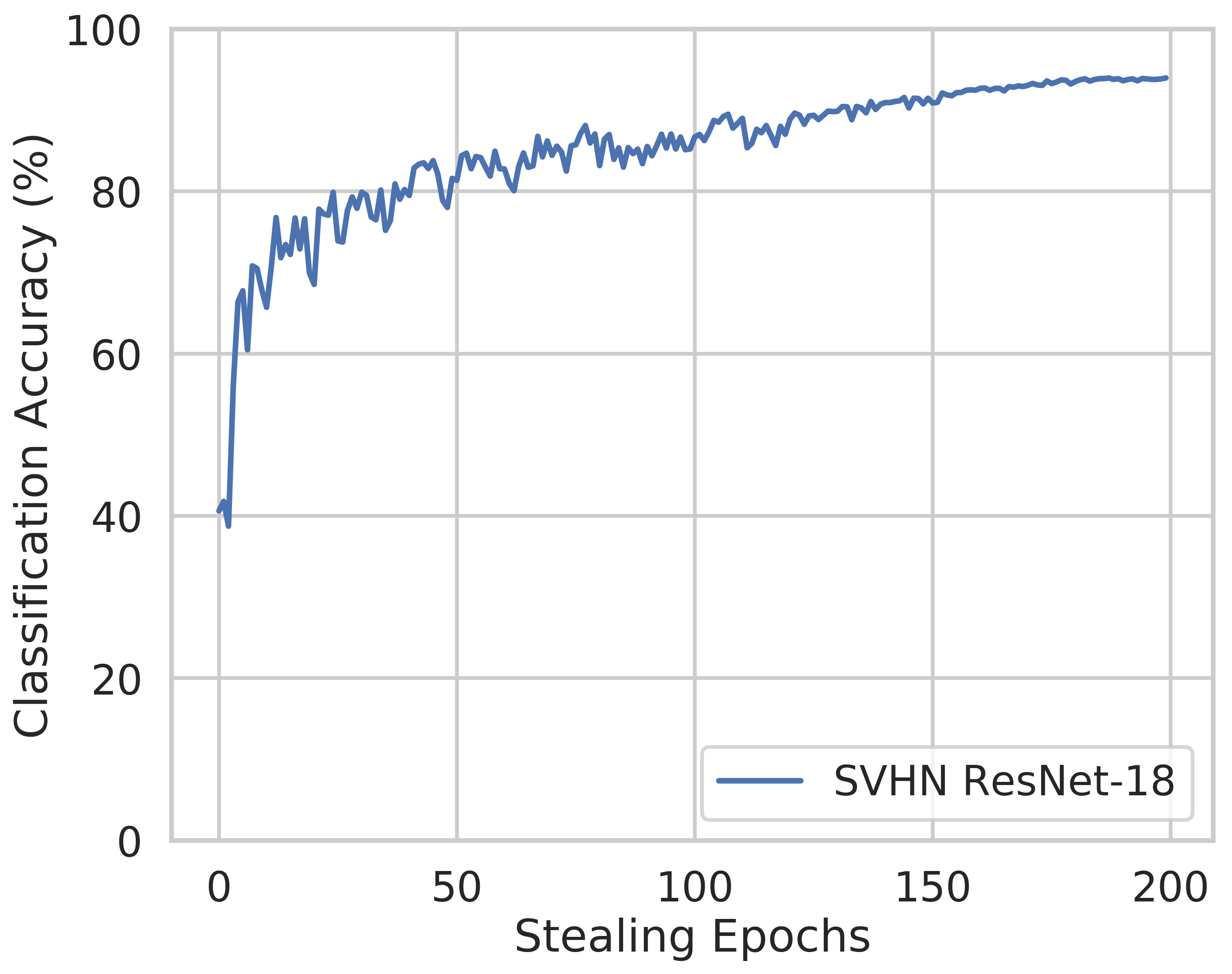}
\caption{ResNet18 on SVHN}
\end{subfigure}
\begin{subfigure}{0.45\linewidth}
\centering
\includegraphics[width=\linewidth]{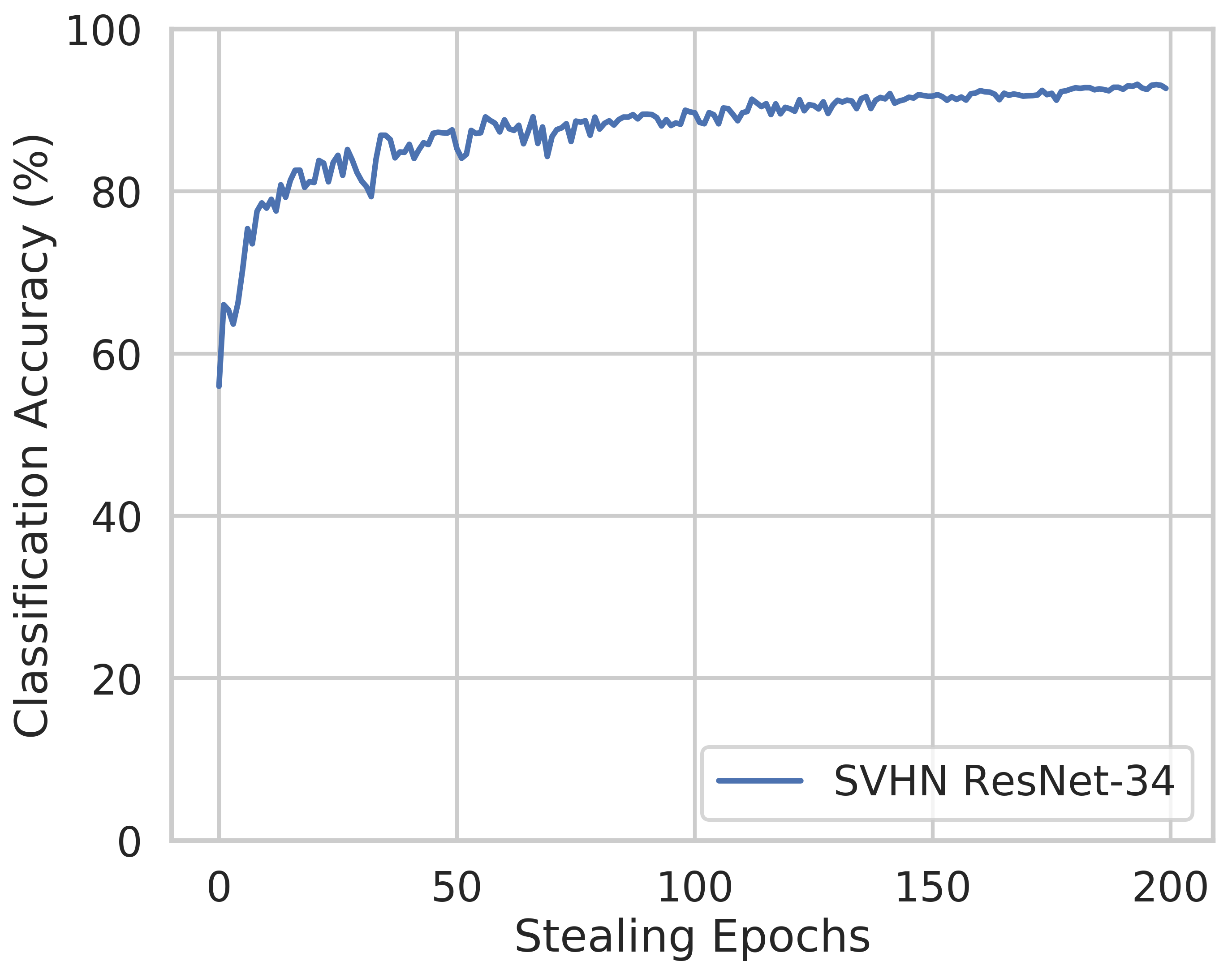}
\caption{ResNet34 on SVHN}
\end{subfigure}
\begin{subfigure}{0.45\linewidth}
\centering
\includegraphics[width=\linewidth]{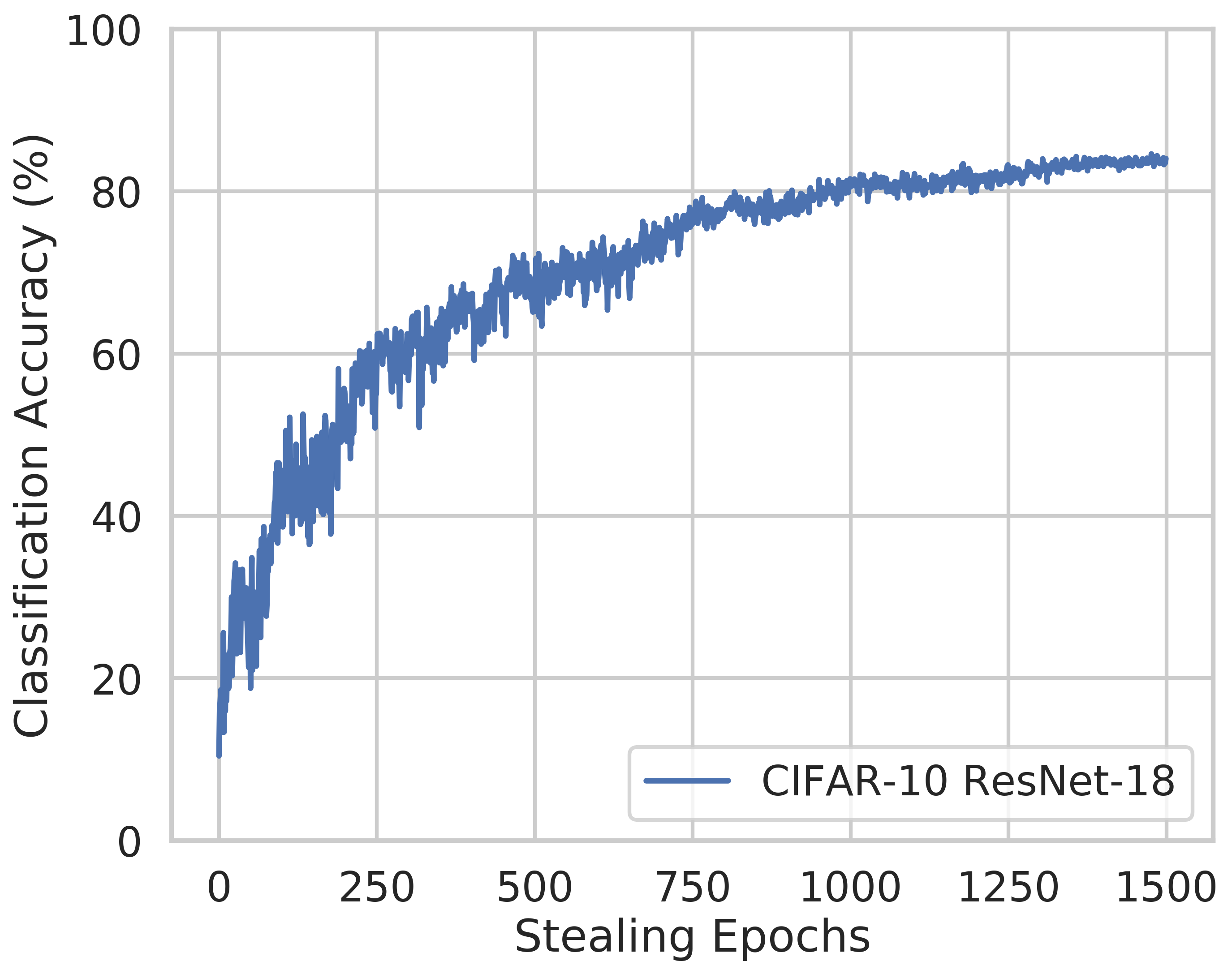}
\caption{ResNet18 on CIFAR10}
\end{subfigure}
\begin{subfigure}{0.45\linewidth}
\centering
\includegraphics[width=\linewidth]{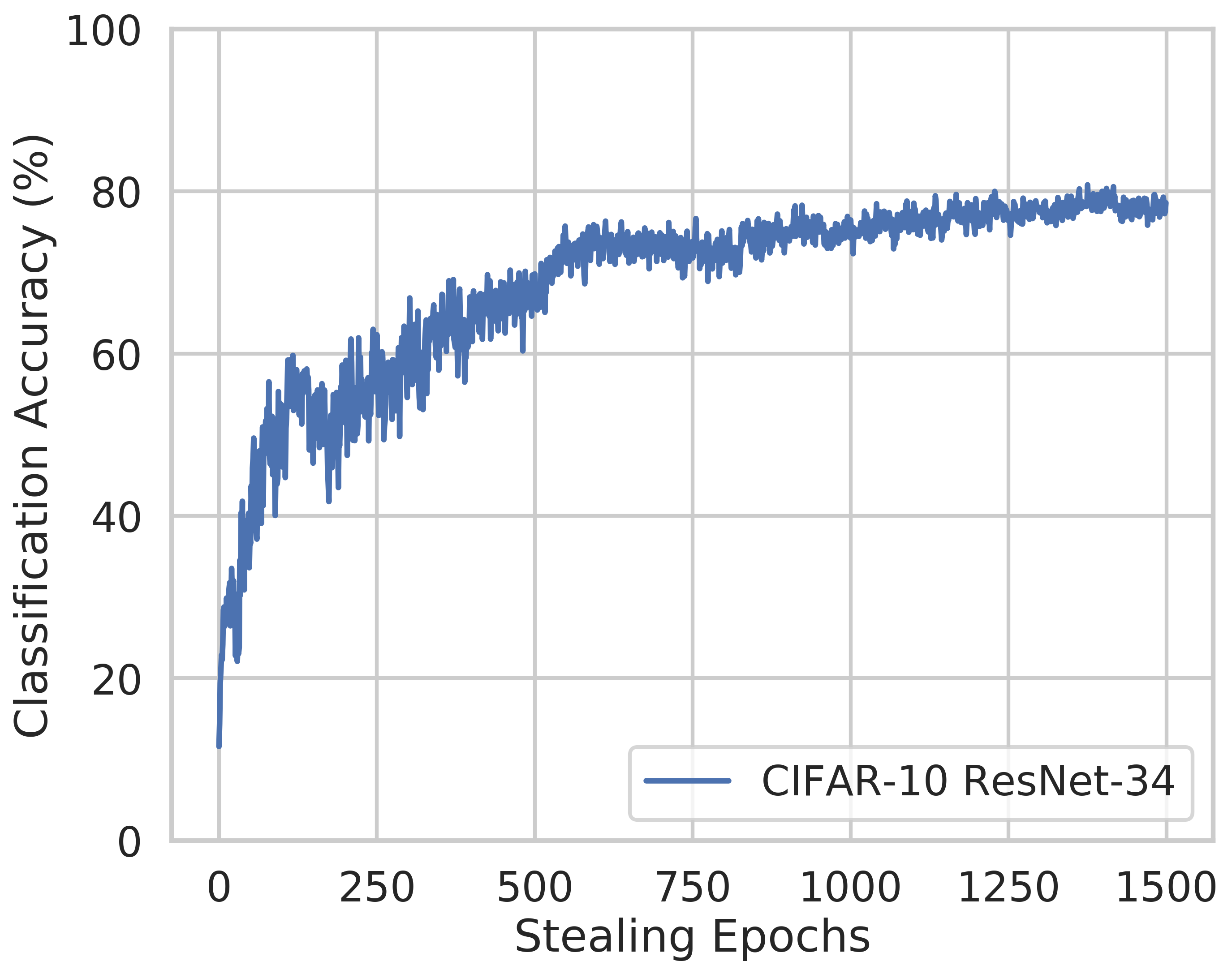}
\caption{ResNet34 on CIFAR10}
\end{subfigure}
\begin{subfigure}{0.45\linewidth}
\centering
\includegraphics[width=\linewidth]{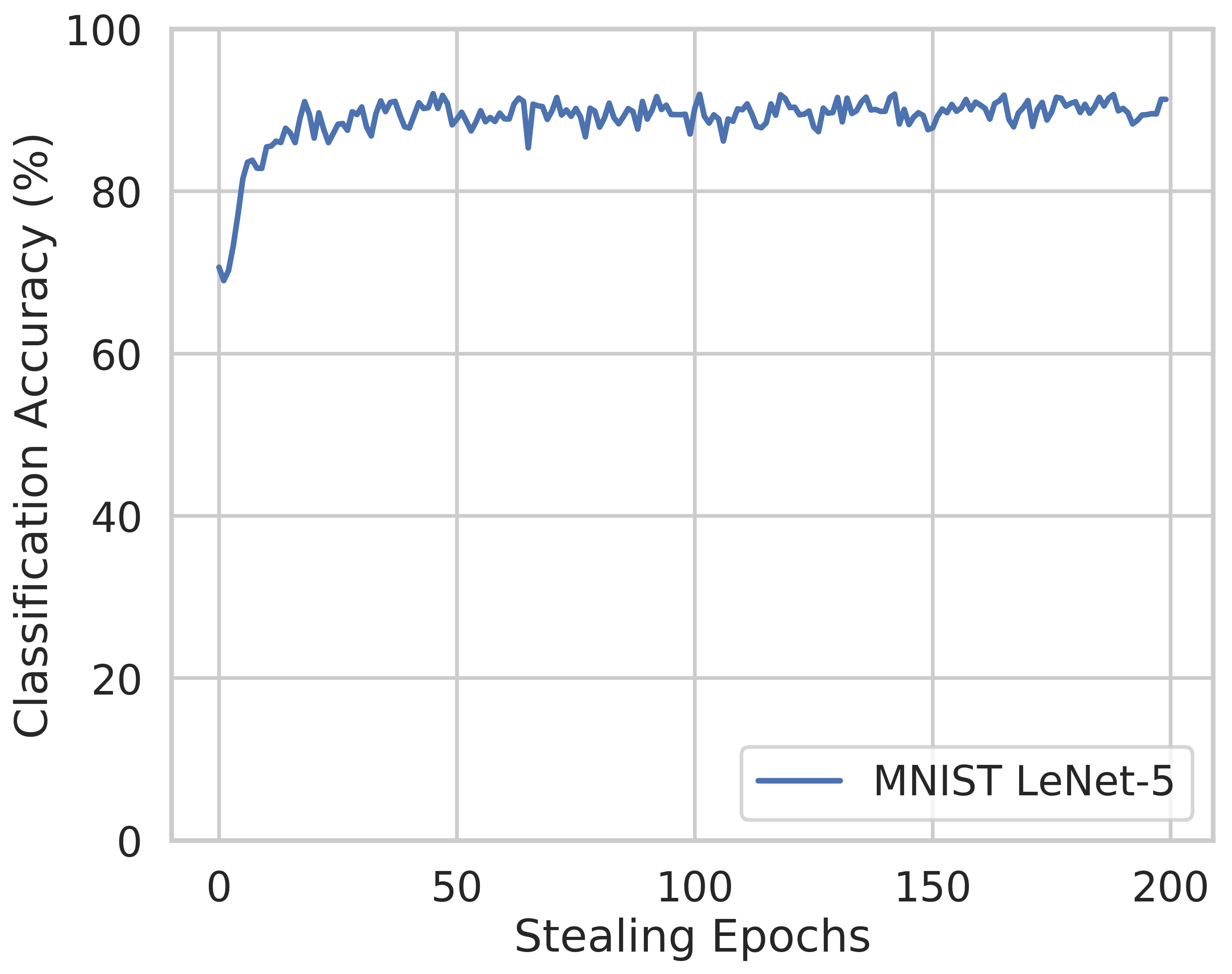}
\caption{LeNet5 on MNIST}
\end{subfigure}
\begin{subfigure}{0.45\linewidth}
\centering
\includegraphics[width=\linewidth]{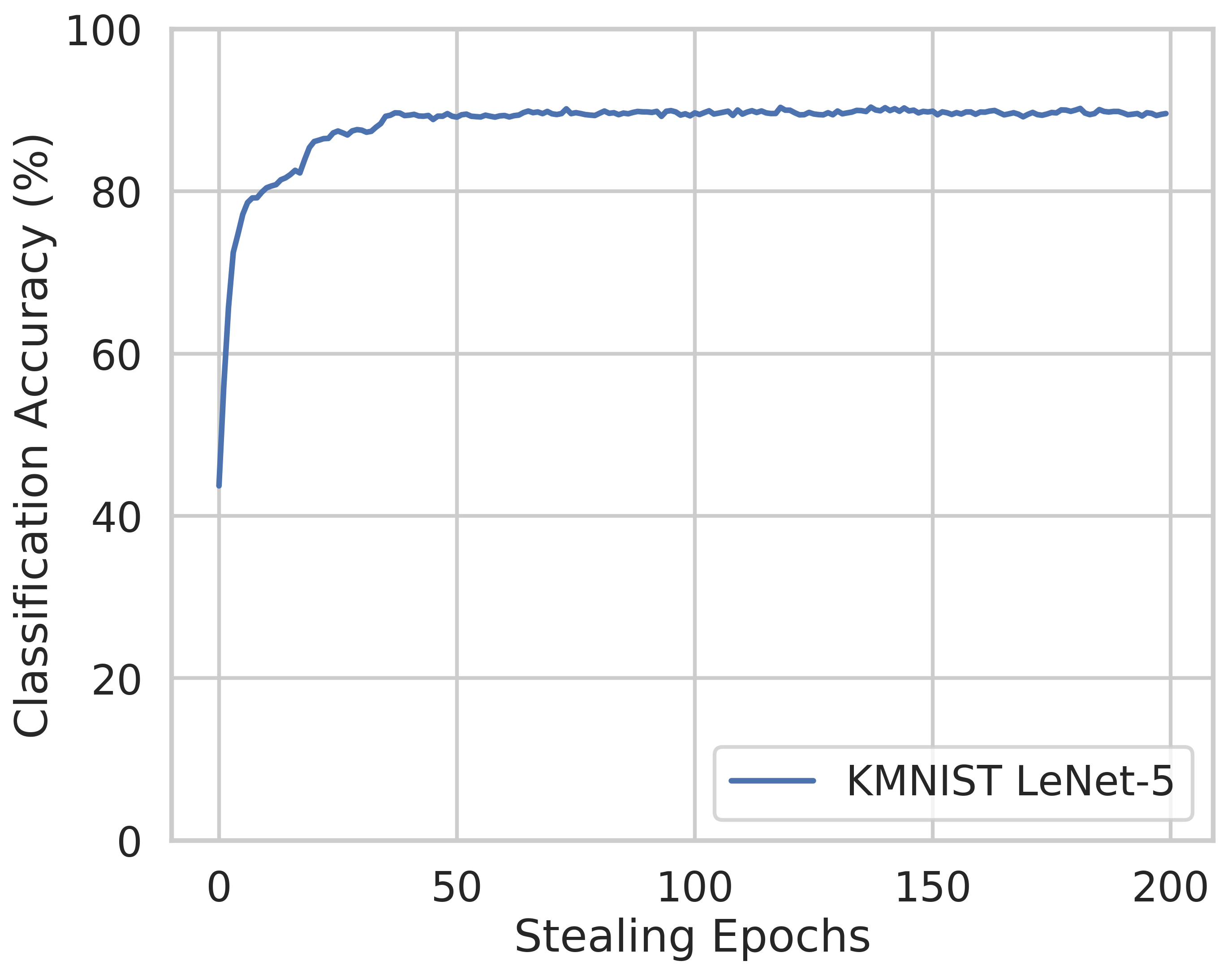}
\caption{LeNet5 on KMNIST}
\end{subfigure}
\vspace{-0.5em}
\caption{\textbf{Substitute model accuracy during attacks.}}
\label{fig:acc}
\vspace{-0.5em}
\end{figure}
Note that we assume that the adversary has no knowledge of any victim's data, which means the adversary cannot evaluate the substitute model on a validation dataset and select the best substitute model during the attack.
If the performance of the stealing attacks fluctuates, then the adversary cannot guarantee the best performance of the substitute model. 
The convergence of the substitute model is essential for stealing attacks without a validation dataset.
Our experiments show that the performance of the substitute model converges after a few stealing epochs (Figure~\ref{fig:acc}).
If the adversary has the knowledge of a validation dataset or the victim's test dataset $\mathcal{D}_{\text{test}}$, the adversary will achieve the best accuracy.
Otherwise, the adversary will use the substitute model in the last stealing epoch ($t=N$). 
We observe the subtle difference between the best accuracy and the last accuracy achieved by the substitute model (0.79\% difference on average for \textit{OPT-SYN} and 1.53\% for \textit{DNN-SYN}).
The stable convergence suggests that our proposed attacks do not rely on a validation dataset.

\xy{We find that model stealing attacks do not require a large query budget in real-world settings. The query to the victim model only occurs in the E-Step, where the adversary uses the synthetic data to get the model prediction. In our experiments, to steal the victim model trained on the MNIST, KMNIST, and SVHN dataset using \textit{OPT-SYN}, the adversary only needs to pay \$30K for all the required queries (around 120M queries) according to the pricing of Amazon AWS~\cite{amazon}.
For for the CIFAR10 dataset, it costs \$187.5K to steal the victim model (around 750M queries). 
The expenses are much less than hiring ML experts and collecting data from scratch. This indicates that the adversary can replicate the function of the existing MLaaS models with much less cost than the victim's actual cost of establishing a machine learning service.
}

\subsection{Sensitivity Analysis of DNN Architectures}
\begin{table}[!tb]
\renewcommand{\arraystretch}{1.3}
\centering
\caption{\eua using Different DNN Architectures.}
\label{tab:arch}
\vspace{-0.5em}
\begin{tabular}{@{}lllcc@{}}
\toprule
Dataset & \begin{tabular}[l]{@{}l@{}} Victim \\ model\end{tabular} & \begin{tabular}[l]{@{}l@{}} Substitute \\ model\end{tabular} & 
\begin{tabular}[c]{@{}c@{}} Victim \\accuracy (\%) \end{tabular} &
\begin{tabular}[c]{@{}c@{}} Substitute \\accuracy (\%) \end{tabular} \\ \midrule
\multirow{2}{*}{MNIST} & ResNet18 & LeNet5 & 99.31 & 97.28 \\ 
& LeNet5 & ResNet18 & 92.03 & 98.13 \\\hline
\multirow{2}{*}{SVHN} & ResNet18 & ResNet34 & 95.40 & 94.64 \\
& ResNet34 & ResNet18 & 95.94 & 94.03 \\ \hline
\multirow{2}{*}{CIFAR10} & ResNet18 & ResNet34 & 91.12 & 82.54 \\
& ResNet34 & ResNet18 & 91.93 & 62.73 \\ \bottomrule
\end{tabular}
\vspace{-0.5em}
\end{table}

\xy{In this section, we consider a more realistic scenario where the adversary has no knowledge of the victim model's architecture. The adversary may choose a different architecture of the substitute model from that of the victim model. \xyy{Thus, we investigate when the adversary uses a different neural network architecture from the victim. 
For the MNIST dataset, we use ResNet18 for the victim model and LeNet5 for the substitute model, and vise versa. For the SVHN and CIFAR10 datasets, we consider ResNet18 and ResNet34 as the victim model and the substitute model's architecture. Because \textit{OPT-SYN} outperforms \textit{DNN-SYN} in most model stealing attacks, we use \textit{OPT-SYN} to evaluate the sensitivity of DNN architectures.

From Table~\ref{tab:arch}, we do not observe a significant performance loss due to different DNN architectures for the MNIST and SVHN dataset, compared with using the same architecture. For the CIFAR10 dataset, we observe subtle performance loss using ResNet34 to steal ResNet18 models. 
The only degradation of performance occurs when the adversary uses a small model ResNet18 to steal a large model ResNet34. }
We believe the degradation is due to the gap between the size of the victim model and the substitute model. }
We find similar performance degradation in many other tasks using knowledge distillation. The performance of the student model (substitute model in our paper) will be degraded if there is a gap between student and teacher (victim)~\cite{mirzadeh2019improved}. 
The adversary can easily avoid performance degradation by selecting a large model. 
From our experiments, if the adversary chooses a DNN model with the same size or the large size compared with the victim model, the adversary will be able to steal a substitute model with high accuracy.

\subsection{Convergence of \eua}
Figure~\ref{fig:acc} illustrates the convergence of \eua. 
We observe that the accuracy of the substitute model always converges at the end of the stealing. 
We observe the subtle difference between the best accuracy and the last accuracy achieved by the substitute model (0.79\% difference on average for \textit{OPT-SYN} and 1.53\% for \textit{DNN-SYN}). 
The stable convergence at the end of the attacks suggests that our proposed attacks do not rely on a test dataset. 
Hence, the adversary can successfully steal the victim model even without knowing test data, which suggests the practicality of \eua and substantially raises the severity of model stealing attacks.

\subsection{Quality Analysis of Synthetic Data}
\label{eval:syn_data}
In this section, we investigate the quality of synthetic data.
Figure~\ref{fig:syn_data} shows examples of the synthetic data used in model stealing. We compare them with the victim's training data. Humans cannot recognize these images, yet our substitute model can be well-trained using these synthetic data. 
\input{synthetic_data}

Therefore, we further investigate synthetic data in terms of quality and diversity. Inspired by the measurements for GANs, we use Inception Score (IS)~\cite{salimans2016improved} and Fr\'{e}chet Inception Distance (FID)~\cite{heusel2017gans} to evaluate the synthetic data. In the experiments, we observe that the synthetic data achieves better quality and higher diversity compared to the auxiliary data.

\textbf{Inception Score (IS)} was originally proposed to measure the quality of generated images using a pre-trained Inception-V3 network. In our experiments, we replaced the Inception-V3 network with the victim models. To be consistent with the concept, we keep Inception Score as the name of our metric. Given the prediction provided by the victim models, Inception Score compares the conditional prediction distribution with the marginal prediction distribution:
\begin{equation}
    \text{IS} = \exp(\mathbb{E}_{\bm{x}} D_{\text{KL}}(p(\bm{y}|\bm{x})) || p(\bm{y})),
\end{equation}
where $D_{\text{KL}}$ denotes the KL divergence. A high Inception Score indicates: 1) generated images are meaningful and predictable to the victim model, so that $p(\bm{y}|\bm{x})$ has low entropy; 2) generated images are diverse, so that $p(\bm{y})$ has high entropy. 

\textbf{Fr\'{e}chet Inception Distance (FID)} was proposed to improve IS by comparing the intermediate features of the victim model. FID models the real images and the synthesis data as two multivariate Gaussian distributions and calculates the distance between the two distributions:
\begin{equation}
    \text{FID} = ||\mu_t - \mu_s||_2^2 + Tr(\Sigma_t + \Sigma_s - 2(\Sigma_t\Sigma_s)^{\frac{1}{2}}),
\end{equation}
where $(\mu_t, \Sigma_t)$ and $(\mu_s, \Sigma_s)$ denote the mean and covariance matrix of intermediate features predicted using training data and synthesis data. $Tr(\cdot)$ denotes the trace of a matrix.
A low FID indicates better image quality and diversity.
FID is shown to be more consistent with human perception than IS~\cite{heusel2017gans} and more robust to mode collapse~\cite{lucic2018gans}. 
In our experiments, we used the features from the layer before the last linear layer and compared our synthesis data with the training data using FID. 
These two metrics are widely used to measure the quality of generated images. 

\begin{table*}[!tb]
\renewcommand{\arraystretch}{1.3}
\centering
\caption{Analysis of Synthetic Data using IS and FID.}
\label{tab:data_eval}
\vspace{-0.5em}
\begin{tabular}{@{}llr|rr|rr|rr@{}}
\toprule
\multirow{2}{*}{Dataset} & \multirow{2}{*}{\shortstack{Model}} & \shortstack{Victim $\mathcal{D}_{\text{train}}$} & \multicolumn{2}{c|}{{Auxiliary $\mathcal{D}_{\text{aux}}$}} & \multicolumn{2}{c|}{{Random $\mathcal{D}_{syn}^{(0)}$}} & \multicolumn{2}{c}{{Synthetic $\mathcal{D}_{syn}^{(N)}$}} \\ \cline{3-9}
 &  & IS & IS & FID & IS & FID & IS & FID \\ \hline
MNIST & LeNet5 & 9.86 & 4.22 & 274.80 & 1.22 & 500.16 & 4.63 & 257.55 \\ 
KMNIST & LeNet5 & 9.96 & 4.75 & 1073.92 & 1.78 & 1498.75 & 4.70 & 962.47 \\ 
CIFAR10 & ResNet18 & 6.89 & 2.58 & 5.34 & 3.45 & 5.82 & 4.32 & 3.31 \\ 
CIFAR10 & ResNet34 & 7.64 & 4.16 & 18.16 & 6.75 & 14.88 & 6.65 & 18.29 \\ 
SVHN & ResNet18 & 6.89 & 2.24 & 7.62 & 3.45 & 5.82 & 4.32 & 3.31 \\ 
SVHN & ResNet34 & 7.64 & 4.16 & 18.16 & 6.75 & 14.88 & 6.65 & 18.29 \\ \bottomrule
\end{tabular}
\vspace{-0.5em}
\end{table*}

We compare the four types of data and report the average values of IS and FID in Table~\ref{tab:data_eval}: 1) victim's training dataset, 2) auxiliary dataset used in the baseline attack, 3) random data generated in the first initialization epoch ($t=0$), 4) the synthetic data generated in the last stealing epoch ($t=N$). 
The value of FID is evaluated by comparing the data with the victim's training data. 
From our analysis, we find that synthetic data usually achieves better quality and high diversity than the auxiliary data (higher IS value and lower FID value). 
On average over six settings, synthetic data $\mathcal{D}_{syn}^{(1)}$ achieves 60.58\% higher IS values and 27.64\% lower FID values than the auxiliary data $\mathcal{D}_{\text{aux}}$, which suggests better quality and higher diversity of synthetic images.
The victim's training data always achieve the highest IS value: the training data is the best representation of the input space among data we investigate. 
The Random Data are always the worst data due to the low IS values and high FID values.


%% file: synthetic_data.tex
\begin{figure*}[!tb]
\centering
\begin{subfigure}{0.1\textwidth}
\centering
\includegraphics[width=0.8\linewidth]{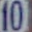}
\end{subfigure}
\begin{subfigure}{0.1\textwidth}
\centering
\includegraphics[width=0.8\linewidth]{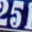}
\end{subfigure}
\begin{subfigure}{0.1\textwidth}
\centering
\includegraphics[width=0.8\linewidth]{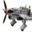}
\end{subfigure}
\begin{subfigure}{0.1\textwidth}
\centering
\includegraphics[width=0.8\linewidth]{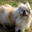}
\end{subfigure}
\begin{subfigure}{0.1\textwidth}
\centering
\includegraphics[width=0.8\linewidth]{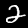}
\end{subfigure}
\begin{subfigure}{0.1\textwidth}
\centering
\includegraphics[width=0.8\linewidth]{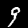}
\end{subfigure}
\begin{subfigure}{0.1\textwidth}
\centering
\includegraphics[width=0.8\linewidth]{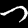}
\end{subfigure}
\begin{subfigure}{0.1\textwidth}
\centering
\includegraphics[width=0.8\linewidth]{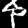}
\end{subfigure}

\par\bigskip
\begin{subfigure}{0.1\textwidth}
\centering
\includegraphics[width=0.8\linewidth]{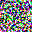}
\end{subfigure}
\begin{subfigure}{0.1\textwidth}
\centering
\includegraphics[width=0.8\linewidth]{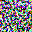}
\end{subfigure}
\begin{subfigure}{0.1\textwidth}
\centering
\includegraphics[width=0.8\linewidth]{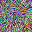}
\end{subfigure}
\begin{subfigure}{0.1\textwidth}
\centering
\includegraphics[width=0.8\linewidth]{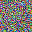}
\end{subfigure}
\begin{subfigure}{0.1\textwidth}
\centering
\includegraphics[width=0.8\linewidth]{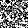}
\end{subfigure}
\begin{subfigure}{0.1\textwidth}
\centering
\includegraphics[width=0.8\linewidth]{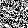}
\end{subfigure}
\begin{subfigure}{0.1\textwidth}
\centering
\includegraphics[width=0.8\linewidth]{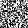}
\end{subfigure}
\begin{subfigure}{0.1\textwidth}
\centering
\includegraphics[width=0.8\linewidth]{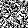}
\end{subfigure}

\par\bigskip
\begin{subfigure}{0.1\textwidth}
\centering
0
\end{subfigure}
\begin{subfigure}{0.1\textwidth}
\centering
5
\end{subfigure}
\begin{subfigure}{0.1\textwidth}
\centering
plane
\end{subfigure}
\begin{subfigure}{0.1\textwidth}
\centering
dog
\end{subfigure}
\begin{subfigure}{0.1\textwidth}
\centering
2
\end{subfigure}
\begin{subfigure}{0.1\textwidth}
\centering
9
\end{subfigure}
\begin{subfigure}{0.1\textwidth}
\centering
\begin{CJK}{UTF8}{min}
つ
\end{CJK}
\end{subfigure}
\begin{subfigure}{0.1\textwidth}
\centering
\begin{CJK}{UTF8}{min}
や
\end{CJK}
\end{subfigure}
\caption{\textbf{Victim's Training Data vs. Synthetic Data.} We synthesize images using \textit{OPT-SYN} with the best substitute model. We compare them with the victim's training data. First row: images in the victim's training dataset. Second row: images in the synthetic dataset. Third Row: corresponding labels of images. Other examples are presented in Appendix (Figure~\ref{fig:svhn_img},~\ref{fig:cifar_img},~\ref{fig:mnist_img},~\ref{fig:kmnist_img}).}
\label{fig:syn_data}
\end{figure*}

%% file: adversarial.tex
DNNs are vulnerable to adversarial examples, a slight modification on the original data sample that can easily fool DNNs~\cite{szegedy2013intriguing,carlini2017towards,madry2017towards,yuan2019adversarial}. 
Black-box adversarial attacks assume that the adversary can only access the output of the model victim instead of its internal information, which is an emerging topic in adversarial attacks.
After \eua, the adversary has full knowledge of the stolen substitute model. Hence, in this section, we try to answer the following question: Can the adversary leverage the knowledge to conduct adversarial attacks against the victim model? 

Transferability is commonly used to conduct black-box adversarial attacks, by training a surrogate model to transfer the adversarial attacks~\cite{papernot2016transferability,liu2016delving}.
In the experiments, we demonstrate how model stealing facilitates black-box adversarial attacks through transferability. 
The evaluation of the black-box adversarial attack is outlined as follows:
1) Steal the victim model using \eua and get a substitute model; 2) Perform a white-box $\ell_{\infty}$-PGD attack against the substitute model and generate adversarial examples; 3) Evaluate the generated adversarial examples on the victim model. 


We evaluate the black-box adversarial attack against the victim model using the substitute model. We implement a white-box $\ell_{\infty}$-PGD attacks~\cite{madry2017towards} and leverage the transferability of adversarial examples to conduct the attack.
PGD attack is an iterative gradient-based attack in a white-box setting and has been proven effective in many machine learning models and tasks. 


In the experiment, we use the test dataset $\mathcal{D}_{test}$ as our evaluation dataset. We follow the adversarial settings in~\cite{madry2017towards} and consider the untargeted adversarial attacks, where adversarial examples can be classified as any classes other than the ground-truth class. 
For the MNIST and KMNIST dataset, we run 40 iterations of $\ell_{\infty}$-PGD attack with a step size of 0.01. We set the maximal perturbation size as $0.3$. For the SVHN and CIFAR10 dataset, we run 20 iterations with a step size of 2/255. The maximal perturbation size is set as $8/255$. 

We report the success rate of adversarial attacks against our substitute model and the victim model (transferring attack) in Table~\ref{tab:adversarial}. We compare the success rate of three adversarial attacks: 1) white-box attacks against the victim model, 2) white-box attacks against the substitute model, and 3) black-box attacks against the victim model via transferring. For the third attack, we evaluate the adversarial examples generated against substitute model (white-box attacks) on the victim model. We show the performance of the white-box $\ell_{\infty}$-PGD attack against the victim model as well. 
\begin{table}[!tb]
\renewcommand{\arraystretch}{1.3}
\centering
\caption{Success rates of adversarial attacks.}
\label{tab:adversarial}
\vspace{-0.5em}
\resizebox{\linewidth}{!}{%
\begin{tabular}{@{}llrrr@{}}
\toprule
\multirow{3}{*}{Dataset} & \multirow{3}{*}{Model}& \multicolumn{3}{c}{Attack success rate (\%)}\\\cline{3-5}
& & \begin{tabular}[r]{@{}r@{}} white-box\\victim model\end{tabular} & \begin{tabular}[c]{@{}c@{}} white-box\\substitute model\end{tabular} & \begin{tabular}[c]{@{}c@{}} black-box\\victim model\end{tabular} \\ \midrule
SVHN& ResNet18 & 99.95 & 99.94 & 98.71 \\ 
SVHN& ResNet34 & 99.93 & 99.90 & 98.21 \\ 
CIFAR10& ResNet18 & 100.00 & 100.00 & 93.60 \\ 
CIFAR10& ResNet34 & 100.00 & 100.00 & 100.00 \\
MNIST& LeNet5 & 86.07 & 99.57 & 92.14 \\
KMNIST& LeNet5 & 66.44 & 99.72 & 98.99 \\ \bottomrule
\end{tabular}
}
\vspace{-1em}
\end{table}
From the experimental results, the black-box adversarial attacks using the substitute model can achieve the same success rate as the white-box, which suggests that the substitute models can transfer the adversarial examples to the victim model successfully. 
Almost all black-box adversarial attacks achieve high accuracy rates (over 90\%). 
We observe that the success rates of the black-box attacks against the victim models are less than that of the white-box attacks against the substitute models, but the change is subtle. Hence, most adversarial examples can be transferred from the substitute model to the victim model. 
Surprisingly, the black-box attacks against the victim model perform even better than the white-box attacks against the victim model on the MNIST and KMNIST dataset. 

%% file: 5-defenses.tex
In this section, we discuss the defense strategies of MLaaS providers and evaluate their effectiveness. Given the good performance of \textit{OPT-SYN} on all the datasets, we evaluate three countermeasures against \textit{OPT-SYN}. We find that the countermeasures are ineffective in defending or detecting proposed \textit{OPT-SYN}.

\subsection{Rounding Prediction}
The MLaaS providers fix the decimals of the output prediction and zero-out the rest to provide only the necessary information. For example, if we round the prediction with 2 decimals, then $round(0.2474, r=2) = 0.25$. 
We deploy rounding predictions with 2 decimals as a defensive strategy. Our experiments show that none of the model stealing attacks are affected by rounding to two decimals (Table~\ref{tab:defense}). 
On average, the best accuracy of the substitute model even increases by 0.55\% and the last accuracy only decreases by 0.30\%.

\begin{table*}[!tb]
\renewcommand{\arraystretch}{1.3}
\centering
\caption{Evaluation of defenses against model stealing.}
\label{tab:defense}
\vspace{-0.5em}
\begin{tabular}{llrrr}
\toprule
Dataset  & \begin{tabular}[l]{@{}l@{}} Model\end{tabular}
&\begin{tabular}[l]{@{}l@{}} Accuracy without defense (\%)\end{tabular}
&\begin{tabular}[l]{@{}l@{}} Accuracy with rounding (\%)\end{tabular}
&\begin{tabular}[l]{@{}l@{}} Accuracy with Top-K (\%)\end{tabular}\\\hline
SVHN     & ResNet34           & 93.19                    & 92.58                          & 88.76\\ 
CIFAR10 & ResNet34           & 80.79                    & 80.31                          & 69.64\\ 
MNIST    & LeNet5            & 92.03                    & 96.69                          & 95.43\\ 
KMNIST   & LeNet5            & 90.37                    & 90.03                          & 91.11\\ \bottomrule
\end{tabular}
\vspace{-0.5em}
\end{table*}

\begin{figure}[!tb]
\centering
\includegraphics[width=0.7\linewidth]{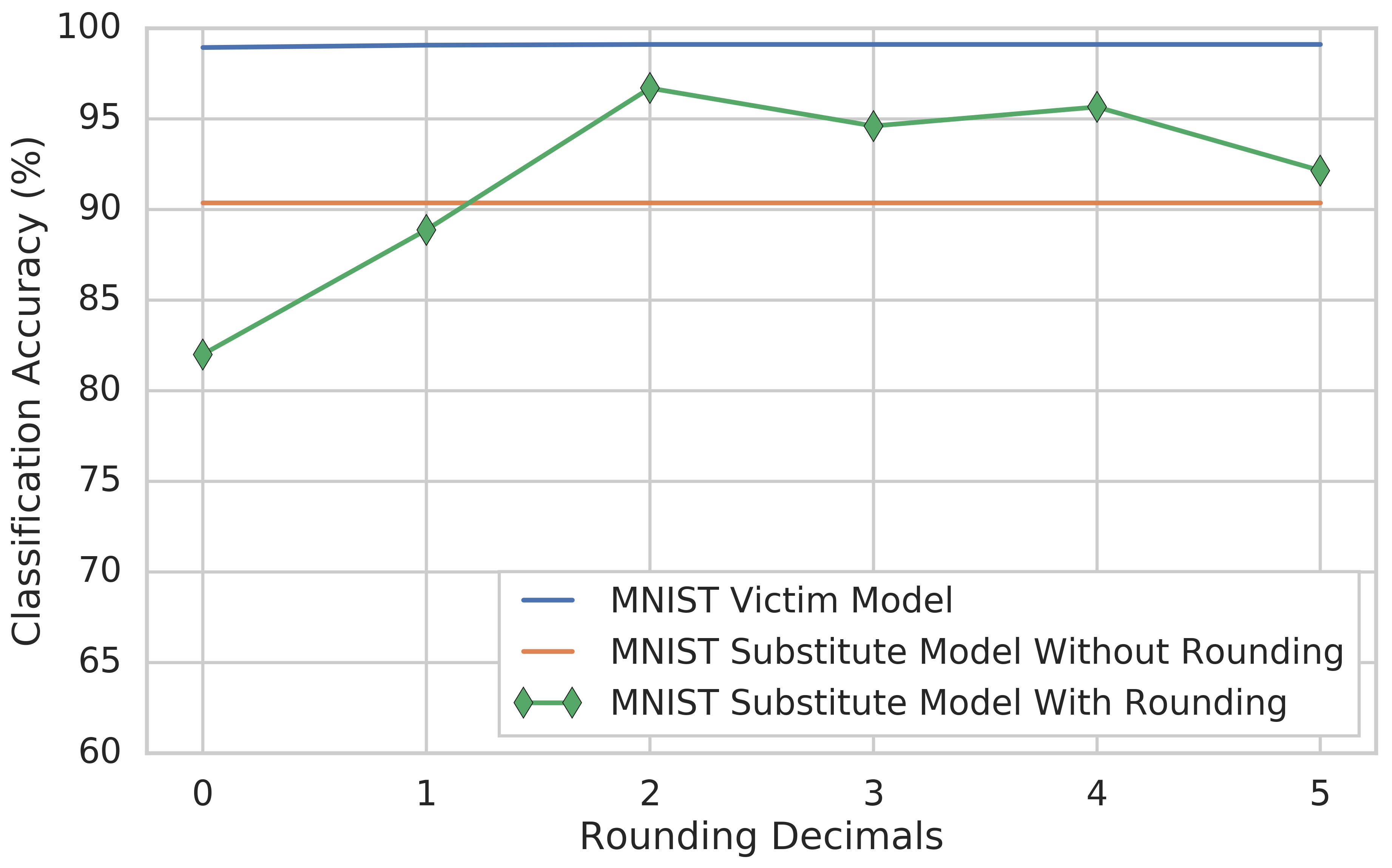}
\caption{\textbf{Evaluation of Rounding Prediction on MNIST.}}
\label{fig:round_mnist}
\end{figure}

\begin{figure}[!tb]
\centering
\includegraphics[width=0.7\linewidth]{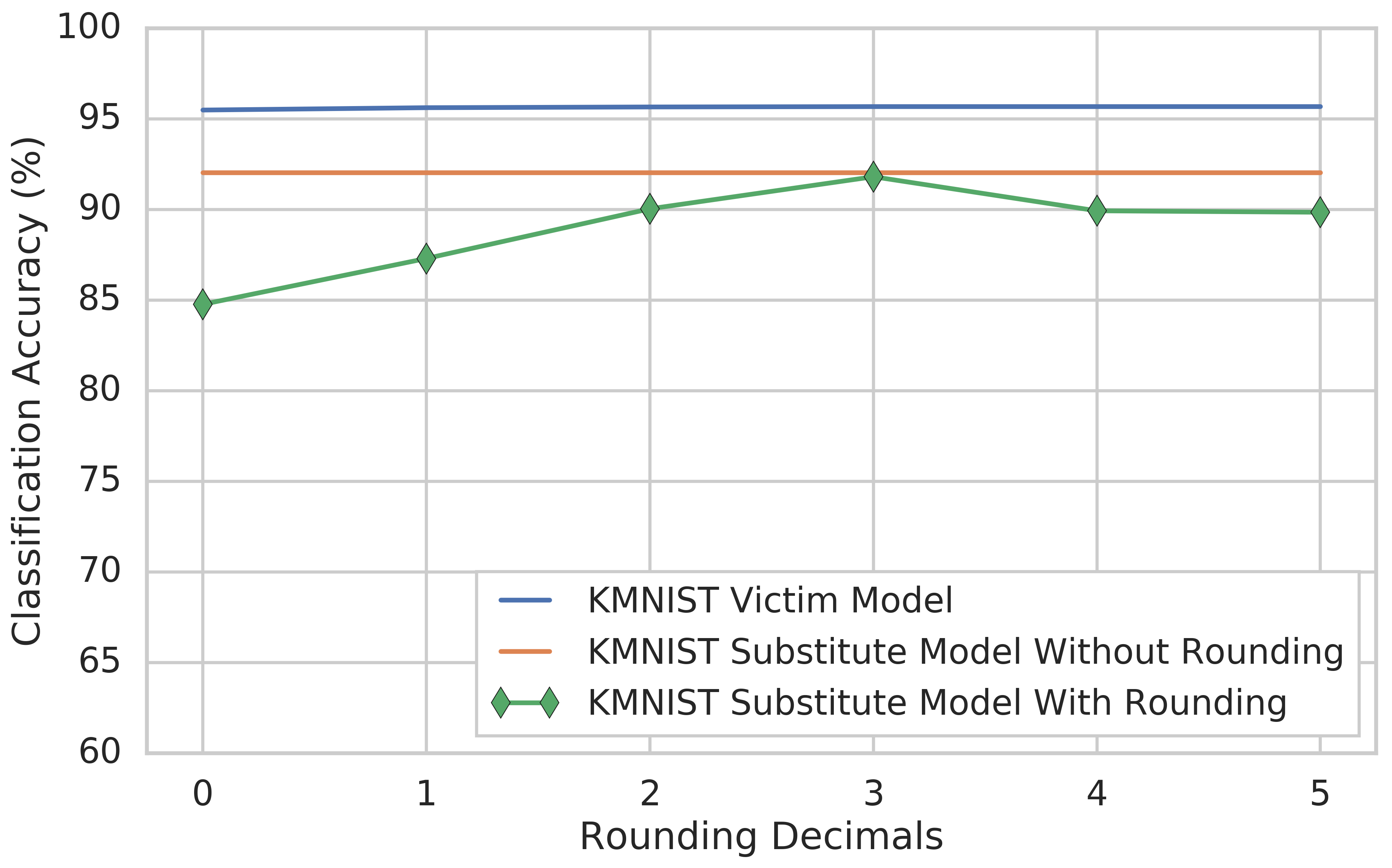}
\caption{\textbf{Evaluation of Rounding Prediction on KMNIST.}}
\label{fig:round_kmnist}
\vspace{-1em}
\end{figure}

We further investigate the impact of rounding decimals on the \eua. 
Figure~\ref{fig:round_mnist} and~\ref{fig:round_kmnist} show the results of experiments with class probabilities rounded to 0-5 decimals. 
We compare the after-rounding classification accuracy of the substitute model and the victim model. 
Class probabilities rounded to 2 to 5 decimals have no effect on the adversary’s success. 
When rounding further to 1 decimal, the attack is weakened, but still successful. 
When we round the precision to 0 decimal - the victim model only outputs $0$ or $1$ - the attack is further weakened, but still can predict in most cases (over 80\% accuracy). 
We observe that rounded information brings the uncertainty of the prediction, while this uncertainty sometimes will slightly improve the training.

\subsection{Top-K Prediction}
Instead of providing the predictions of all the classes, MLaaS providers can provide partial information - predictions of K classes with the highest probabilities. 
Since this defense can be easily detected by the adversary, we assume the adversary is aware of the top-K defenses equipped by the MLaaS provider and will try to circumvent such defense. 
Therefore, the adversary can slightly change the attack by making up the missing probabilities of the rest classes. The adversary remains the probabilities of the Top-K classes and fills up the rest classes with the same probabilities.
For example, given an prediction output of $[0.5, 0.02, 0.3, 0.02, 0.15, 0.01]$, by using Top-2 defense, MLaaS provider can hide the predictions of eight classes and only respond with the prediction of $[0.5, 0.0, 0.3, 0.0, 0.0, 0.0]$. By knowing the top-k defense, The adversary can then convert the predictions to $[0.5, 0.05, 0.3, 0.05, 0.05, 0.05]$ and resume \eua.

\begin{figure}[!tb]
\centering
\includegraphics[width=0.7\linewidth]{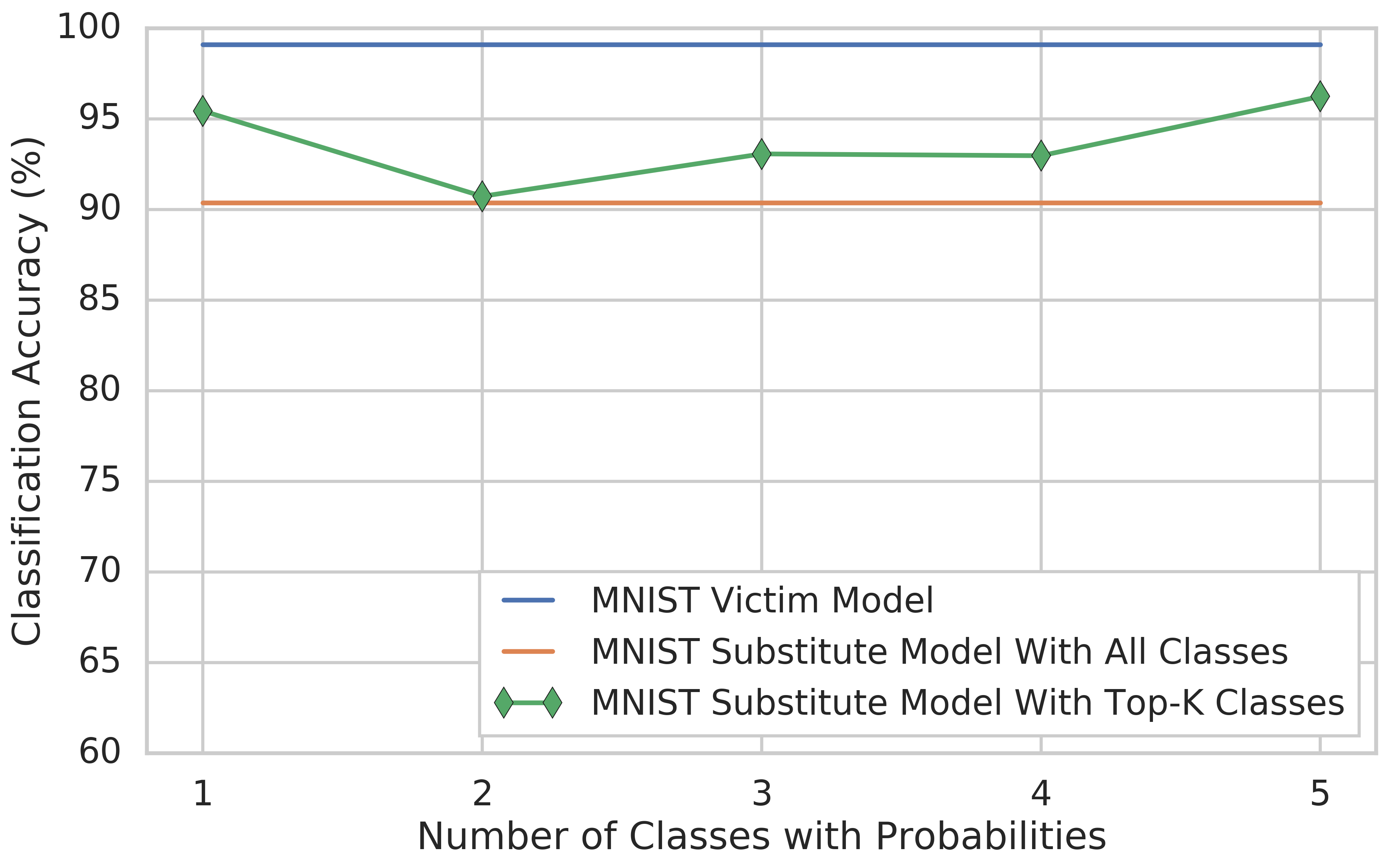}
\caption{\textbf{Evaluation of Top-K Prediction on MNIST.}}
\label{fig:topk_mnist}
\vspace{-0.2cm}
\end{figure}

\begin{figure}[!tb]
\centering
\includegraphics[width=0.7\linewidth]{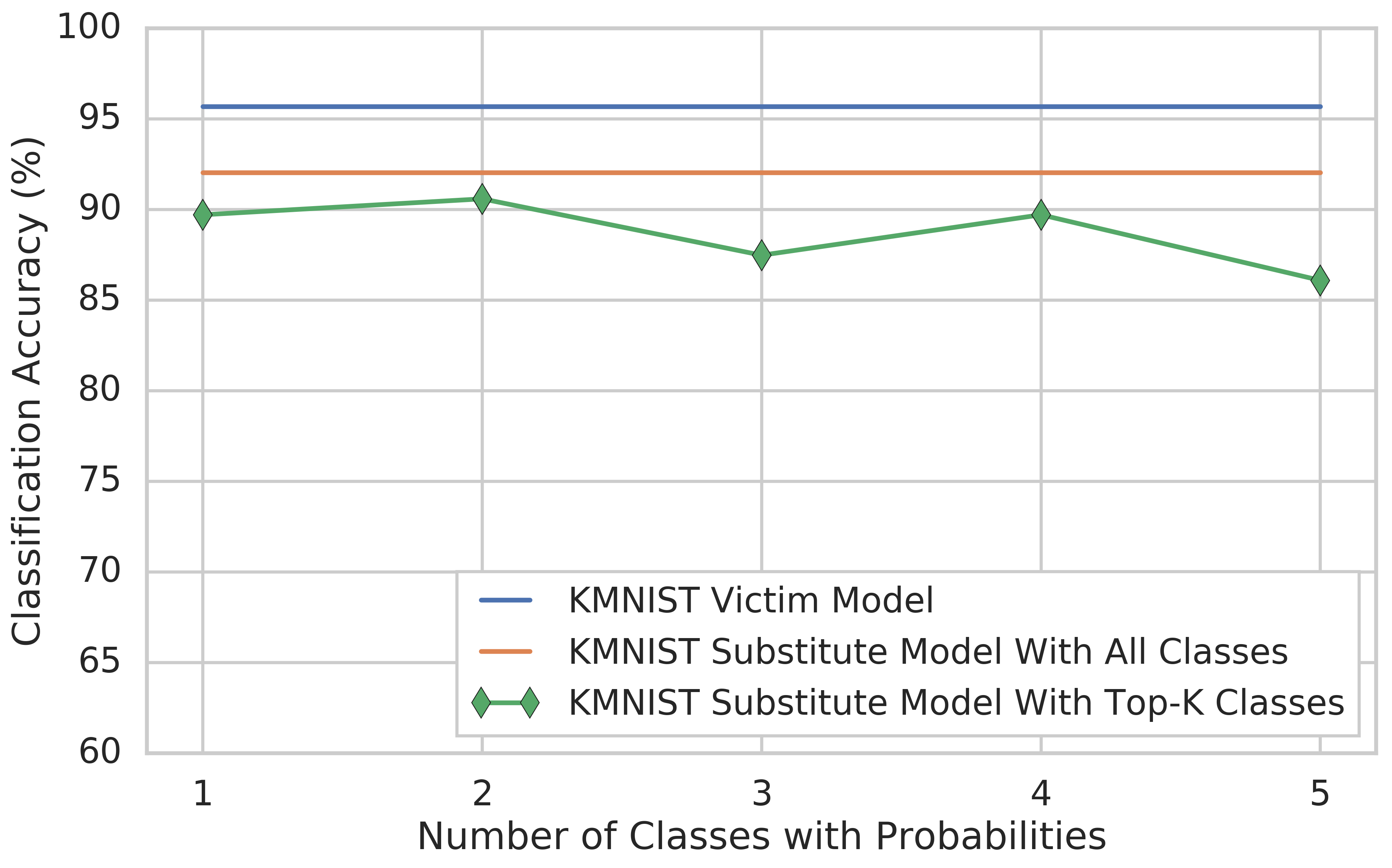}
\caption{\textbf{Evaluation of Top-K Prediction on KMNIST.}}
\label{fig:topk_kmnist}
\vspace{-1em}
\end{figure}

From the experiments, we observe that Top-1 prediction will not affect much on most datasets (Table~\ref{tab:defense}). 
For the MNIST and KMNIST datasets, we find that the accuracy of the substitute model even gets improved. Top-1 prediction is only effective in preventing model stealing on the CIFAR10 dataset.
However, we believe Top-1 prediction is a very strong defense, which will also affect normal users by providing very limited information. On average, the best accuracy of the substitute model with Top-1 prediction is only decreased by 2.86\%.
In addition, we investigate the impact of probabilities with different numbers ($K$) of classes on our model stealing attacks (Figure~\ref{fig:topk_mnist} and~\ref{fig:topk_kmnist}). 
The performance of model stealing attacks is not decreased with fewer classes providing probabilities (small $K$).
We find our attacks are minimally impacted by reducing the informativeness of black-box predictions in the response.



\subsection{Anomaly Detection}
Anomaly detection identifies the abnormal queries sent from users and detects the abnormal behavior that deviates from normal ones. 
For example, PRADA assumes that the distance between normal queries follows Gaussian distribution and detects the abnormal queries~\cite{juuti2018prada}. 
By evaluating how deviated the distance from Gaussian distribution, PRADA detects model stealing attacks. For the details of PRADA, we refer readers to~\cite{juuti2018prada}.
We evaluate the effectiveness of anomaly detection using PRADA against \textit{OPT-SYN}
We analyzed 300,000 image samples from the first five stealing epochs on the MNIST dataset. 
None of the synthetic images can be detected by PRADA.
We believe that because the images are generated starting from the Gaussian distribution, the distances between queried images are too small to be detected by PRADA.
Moreover, we find it is not practical for MLaaS providers to deploy a PRADA detector due to its high response time. 
In our experiments, it takes about 33 hours to process 300,000 images (2.46 images per second on average).
With more images to be detected, the average response time will be further increased.
Therefore, PRADA is ineffective and infeasible to detect the proposed \eua.

%% file: 6-related.tex
\subsection{Model Stealing Attacks}
Several studies have been proposed for model stealing attacks. Tramer \etal investigated stealing model parameters using equation solving~\cite{tramer2016stealing}. However, this approach is hard to extend to DNNs, which contains a larger number of than conventional machine learning models do. 
Papernot \etal proposed a similar framework to steal DNNs by training a substitute model~\cite{papernot2017practical}. Their goal is to approximate the victim model's decision boundaries to facilitate the adversarial attacks rather than to maximize the substitute model's accuracy. Thus their substitute model achieves a much lower classification accuracy compared to our work. In addition, to generate adversarial examples, their approach requires a small set of inputs that represents the input domain. 
\xyy{In our work, we eliminate this requirement, where the adversary does not need to have prior knowledge, making the attacks more feasible in the real world.}
From the experimental results, the stolen model from \eua achieves a higher accuracy compared to that from ~\cite{papernot2017practical}.
Existing model stealing attacks against DNNs require an auxiliary dataset. 
Orekondy \etal proposed stealing attacks that assume access to a large dataset and use active learning to select the best samples to query~\cite{orekondy2019knockoff}. Correia-Silva \etal leveraged public datasets from the same task domain but with different distributions to steal DNNs.
\xy{Different from these works, we assume the adversary does not have any auxiliary data related to the task domain. 
The experiments show that \eua can achieve comparable performance, compared with the attacks using auxiliary datasets.}

\xy{
Zhou~\etal used the same assumption with our work - unknown victim's training data and leveraged a generative model to synthesize the training data~\cite{zhou2020dast}. However, the goal of DaST is to perform a successful adversarial attack, which is different from ours. In this paper, we aim to improve the prediction performance of the substitute model. Accordingly, the substitute model trained by DaST achieves much lower accuracy than \eua.
}
\xy{
MAZE investigated a similar problem with our work, namely data-free model stealing attack~\cite{kariyappa2021maze}. To address the problem of unknown victim's training data, MAZE tried to solve the same challenge as our work, that is in generating synthetic data, the gradient for updating the synthetic data cannot be backpropagated using the victim model. MAZE addressed this issue by approximating the gradients from the victim model using zeroth-order gradient estimation, which is widely used in black-box adversarial attacks, whereas in our work, we generate the gradients by using the substitute model as a proxy of the victim model.
Both approaches achieve comparable attacking performance. In addition, the two approaches are orthogonal and could be further integrated together for better performance. We will explore the new approach benefiting from \eua and MAZE in the future. 
}

\vspace{-0.5em}
\subsection{Model Stealing Defenses}
Several detection approaches have been proposed for model stealing attacks. Juuti \etal detected the deviated distribution of queries from normal behavior~\cite{juuti2018prada}.
Similarly, Kesarwani \etal proposed a detection tool that uses information gain to measure the model learning rate by users with the increasing number of queries~\cite{kesarwani2018model}. The learning rate is measured to the coverage of the input feature space in the presence of collusion. 
We evaluate \cite{juuti2018prada} in our experiments and find that the detection approach is ineffective for our model stealing attacks.


\subsection{Knowledge Distillation}
\label{sec:related_knowledge}
In our model stealing attacks, we use distillation to transfer knowledge from the victim model to the substitute model.
Knowledge distillation is widely used in model compression by transferring the knowledge from one model (teacher model) to another (student model)~\cite{hinton2015distilling,bucilua2006model}. 
Most knowledge distillation approaches require the knowledge of training data.
Recently, knowledge distillation without training data has recently been investigated~\cite{chen2019data,lopes2017data,nayak2019zero,yin2020dreaming,haroush2020knowledge} when the training data is infeasible due to large data size or privacy concerns. 
\xy{However, these data-free knowledge distillation approaches cannot be used for model stealing since the adversary cannot acquire the required information.
For example, the model gradient  is required to update the parameters of the generator in~\cite{chen2019data}. Similarly, model parameters are required to calculate class similarity in~\cite{nayak2019zero} or to calculate the feature map statistics in~\cite{yin2020dreaming} and batch normalization statistics in~\cite{haroush2020knowledge}.
Therefore, beyond data-free knowledge distillation, we introduce two data synthesis approaches in \eua to construct a novel data-free model stealing attack.
}




%% file: 8-conclusion.tex
We demonstrated that our attacks successfully stole various DNNs from the MLaaS providers without any data hurdles. 
Even without the knowledge on the victim's dataset, \eua outperforms the two baseline attacks by 44.57\%  on average of four datasets in terms of best accuracy. 
Our experimental results illustrated the better quality and higher diversity of the generated synthetic data compared with the auxiliary data, which benefits \eua.
In addition, most existing defenses are ineffective to prevent \eua, where new countermeasures should be provided.
Moreover, the stolen model can be used to conduct black-box adversarial attacks against the victim model, and sometimes the black-box attack achieves higher success rates compared with the white-box attack. 
\xyy{In this paper, we only target image classification tasks and small datasets. We will extend our work to other machine learning tasks and more complex datasets such as image segmentation and ImageNet in the future.}

\xy{Further, this paper investigated a critical challenge in the existing machine learning services. By successfully launching the model stealing attack, the adversary can provide the same machine learning service with a much lower price compared to MLaaS provider due to the low cost of model stealing attacks.
Moreover, model stealing attacks may facilitate further serious attacks that have been found in recent machine learning security research.
We hope the severity of model stealing attacks can attract the attention of the community and encourage researchers in academia and industry to investigate effective countermeasures.}

%

%% file: 9-appendix.tex
\section*{Synthetic Example Images by \textit{OPT-SYN}}
\begin{figure*}[!b]
\centering
\begin{subfigure}{0.09\textwidth}
\centering
\includegraphics[width=0.9\linewidth]{figs/imgs/SVHN_train_0.png}
\end{subfigure}
\begin{subfigure}{0.09\textwidth}
\centering
\includegraphics[width=0.9\linewidth]{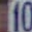}
\end{subfigure}
\begin{subfigure}{0.09\textwidth}
\centering
\includegraphics[width=0.9\linewidth]{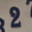}
\end{subfigure}
\begin{subfigure}{0.09\textwidth}
\centering
\includegraphics[width=0.9\linewidth]{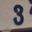}
\end{subfigure}
\begin{subfigure}{0.09\textwidth}
\centering
\includegraphics[width=0.9\linewidth]{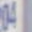}
\end{subfigure}
\begin{subfigure}{0.09\textwidth}
\centering
\includegraphics[width=0.9\linewidth]{figs/imgs/SVHN_train_5.png}
\end{subfigure}
\begin{subfigure}{0.09\textwidth}
\centering
\includegraphics[width=0.9\linewidth]{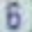}
\end{subfigure}
\begin{subfigure}{0.09\textwidth}
\centering
\includegraphics[width=0.9\linewidth]{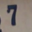}
\end{subfigure}
\begin{subfigure}{0.09\textwidth}
\centering
\includegraphics[width=0.9\linewidth]{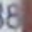}
\end{subfigure}
\begin{subfigure}{0.09\textwidth}
\centering
\includegraphics[width=0.9\linewidth]{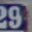}
\end{subfigure}

\par\bigskip
\begin{subfigure}{0.09\textwidth}
\centering
\includegraphics[width=0.9\linewidth]{figs/imgs/SVHN_syn_0.png}
\end{subfigure}
\begin{subfigure}{0.09\textwidth}
\centering
\includegraphics[width=0.9\linewidth]{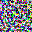}
\end{subfigure}
\begin{subfigure}{0.09\textwidth}
\centering
\includegraphics[width=0.9\linewidth]{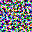}
\end{subfigure}
\begin{subfigure}{0.09\textwidth}
\centering
\includegraphics[width=0.9\linewidth]{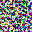}
\end{subfigure}
\begin{subfigure}{0.09\textwidth}
\centering
\includegraphics[width=0.9\linewidth]{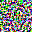}
\end{subfigure}
\begin{subfigure}{0.09\textwidth}
\centering
\includegraphics[width=0.9\linewidth]{figs/imgs/SVHN_syn_5.png}
\end{subfigure}
\begin{subfigure}{0.09\textwidth}
\centering
\includegraphics[width=0.9\linewidth]{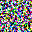}
\end{subfigure}
\begin{subfigure}{0.09\textwidth}
\centering
\includegraphics[width=0.9\linewidth]{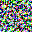}
\end{subfigure}
\begin{subfigure}{0.09\textwidth}
\centering
\includegraphics[width=0.9\linewidth]{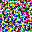}
\end{subfigure}
\begin{subfigure}{0.09\textwidth}
\centering
\includegraphics[width=0.9\linewidth]{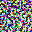}
\end{subfigure}

\par\bigskip
\begin{subfigure}{0.09\textwidth}
\centering
0
\end{subfigure}
\begin{subfigure}{0.09\textwidth}
\centering
1
\end{subfigure}
\begin{subfigure}{0.09\textwidth}
\centering
2
\end{subfigure}
\begin{subfigure}{0.09\textwidth}
\centering
3
\end{subfigure}
\begin{subfigure}{0.09\textwidth}
\centering
4
\end{subfigure}
\begin{subfigure}{0.09\textwidth}
\centering
5
\end{subfigure}
\begin{subfigure}{0.09\textwidth}
\centering
6
\end{subfigure}
\begin{subfigure}{0.09\textwidth}
\centering
7
\end{subfigure}
\begin{subfigure}{0.09\textwidth}
\centering
8
\end{subfigure}
\begin{subfigure}{0.09\textwidth}
\centering
9
\end{subfigure}
\caption{\textbf{The SVHN Dataset}}
\label{fig:svhn_img}
\end{figure*}

\begin{figure*}[!tb]
\centering
\begin{subfigure}{0.09\textwidth}
\centering
\includegraphics[width=0.9\linewidth]{figs/imgs/CIFAR_train_0.png}
\end{subfigure}
\begin{subfigure}{0.09\textwidth}
\centering
\includegraphics[width=0.9\linewidth]{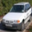}
\end{subfigure}
\begin{subfigure}{0.09\textwidth}
\centering
\includegraphics[width=0.9\linewidth]{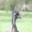}
\end{subfigure}
\begin{subfigure}{0.09\textwidth}
\centering
\includegraphics[width=0.9\linewidth]{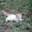}
\end{subfigure}
\begin{subfigure}{0.09\textwidth}
\centering
\includegraphics[width=0.9\linewidth]{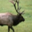}
\end{subfigure}
\begin{subfigure}{0.09\textwidth}
\centering
\includegraphics[width=0.9\linewidth]{figs/imgs/CIFAR_train_5.png}
\end{subfigure}
\begin{subfigure}{0.09\textwidth}
\centering
\includegraphics[width=0.9\linewidth]{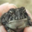}
\end{subfigure}
\begin{subfigure}{0.09\textwidth}
\centering
\includegraphics[width=0.9\linewidth]{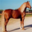}
\end{subfigure}
\begin{subfigure}{0.09\textwidth}
\centering
\includegraphics[width=0.9\linewidth]{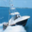}
\end{subfigure}
\begin{subfigure}{0.09\textwidth}
\centering
\includegraphics[width=0.9\linewidth]{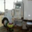}
\end{subfigure}

\par\bigskip
\begin{subfigure}{0.09\textwidth}
\centering
\includegraphics[width=0.9\linewidth]{figs/imgs/CIFAR_syn_0.png}
\end{subfigure}
\begin{subfigure}{0.09\textwidth}
\centering
\includegraphics[width=0.9\linewidth]{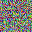}
\end{subfigure}
\begin{subfigure}{0.09\textwidth}
\centering
\includegraphics[width=0.9\linewidth]{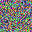}
\end{subfigure}
\begin{subfigure}{0.09\textwidth}
\centering
\includegraphics[width=0.9\linewidth]{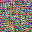}
\end{subfigure}
\begin{subfigure}{0.09\textwidth}
\centering
\includegraphics[width=0.9\linewidth]{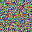}
\end{subfigure}
\begin{subfigure}{0.09\textwidth}
\centering
\includegraphics[width=0.9\linewidth]{figs/imgs/CIFAR_syn_5.png}
\end{subfigure}
\begin{subfigure}{0.09\textwidth}
\centering
\includegraphics[width=0.9\linewidth]{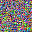}
\end{subfigure}
\begin{subfigure}{0.09\textwidth}
\centering
\includegraphics[width=0.9\linewidth]{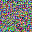}
\end{subfigure}
\begin{subfigure}{0.09\textwidth}
\centering
\includegraphics[width=0.9\linewidth]{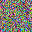}
\end{subfigure}
\begin{subfigure}{0.09\textwidth}
\centering
\includegraphics[width=0.9\linewidth]{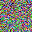}
\end{subfigure}

\par\bigskip
\begin{subfigure}{0.09\textwidth}
\centering
plane
\end{subfigure}
\begin{subfigure}{0.09\textwidth}
\centering
car
\end{subfigure}
\begin{subfigure}{0.09\textwidth}
\centering
bird
\end{subfigure}
\begin{subfigure}{0.09\textwidth}
\centering
cat
\end{subfigure}
\begin{subfigure}{0.09\textwidth}
\centering
deer
\end{subfigure}
\begin{subfigure}{0.09\textwidth}
\centering
dog
\end{subfigure}
\begin{subfigure}{0.09\textwidth}
\centering
frog
\end{subfigure}
\begin{subfigure}{0.09\textwidth}
\centering
horse
\end{subfigure}
\begin{subfigure}{0.09\textwidth}
\centering
ship
\end{subfigure}
\begin{subfigure}{0.09\textwidth}
\centering
truck
\end{subfigure}
\caption{\textbf{The CIFAR-10 Dataset}}
\label{fig:cifar_img}
\end{figure*}

\begin{figure*}[!tb]
\centering
\begin{subfigure}{0.09\textwidth}
\centering
\includegraphics[width=0.9\linewidth]{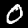}
\end{subfigure}
\begin{subfigure}{0.09\textwidth}
\centering
\includegraphics[width=0.9\linewidth]{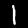}
\end{subfigure}
\begin{subfigure}{0.09\textwidth}
\centering
\includegraphics[width=0.9\linewidth]{figs/imgs/MNIST_train_2.png}
\end{subfigure}
\begin{subfigure}{0.09\textwidth}
\centering
\includegraphics[width=0.9\linewidth]{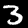}
\end{subfigure}
\begin{subfigure}{0.09\textwidth}
\centering
\includegraphics[width=0.9\linewidth]{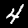}
\end{subfigure}
\begin{subfigure}{0.09\textwidth}
\centering
\includegraphics[width=0.9\linewidth]{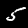}
\end{subfigure}
\begin{subfigure}{0.09\textwidth}
\centering
\includegraphics[width=0.9\linewidth]{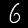}
\end{subfigure}
\begin{subfigure}{0.09\textwidth}
\centering
\includegraphics[width=0.9\linewidth]{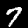}
\end{subfigure}
\begin{subfigure}{0.09\textwidth}
\centering
\includegraphics[width=0.9\linewidth]{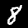}
\end{subfigure}
\begin{subfigure}{0.09\textwidth}
\centering
\includegraphics[width=0.9\linewidth]{figs/imgs/MNIST_train_9.png}
\end{subfigure}

\par\bigskip
\begin{subfigure}{0.09\textwidth}
\centering
\includegraphics[width=0.9\linewidth]{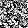}
\end{subfigure}
\begin{subfigure}{0.09\textwidth}
\centering
\includegraphics[width=0.9\linewidth]{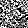}
\end{subfigure}
\begin{subfigure}{0.09\textwidth}
\centering
\includegraphics[width=0.9\linewidth]{figs/imgs/MNIST_syn_2.png}
\end{subfigure}
\begin{subfigure}{0.09\textwidth}
\centering
\includegraphics[width=0.9\linewidth]{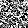}
\end{subfigure}
\begin{subfigure}{0.09\textwidth}
\centering
\includegraphics[width=0.9\linewidth]{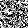}
\end{subfigure}
\begin{subfigure}{0.09\textwidth}
\centering
\includegraphics[width=0.9\linewidth]{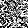}
\end{subfigure}
\begin{subfigure}{0.09\textwidth}
\centering
\includegraphics[width=0.9\linewidth]{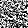}
\end{subfigure}
\begin{subfigure}{0.09\textwidth}
\centering
\includegraphics[width=0.9\linewidth]{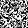}
\end{subfigure}
\begin{subfigure}{0.09\textwidth}
\centering
\includegraphics[width=0.9\linewidth]{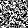}
\end{subfigure}
\begin{subfigure}{0.09\textwidth}
\centering
\includegraphics[width=0.9\linewidth]{figs/imgs/MNIST_syn_9.png}
\end{subfigure}

\par\bigskip
\begin{subfigure}{0.09\textwidth}
\centering
0
\end{subfigure}
\begin{subfigure}{0.09\textwidth}
\centering
1
\end{subfigure}
\begin{subfigure}{0.09\textwidth}
\centering
2
\end{subfigure}
\begin{subfigure}{0.09\textwidth}
\centering
3
\end{subfigure}
\begin{subfigure}{0.09\textwidth}
\centering
4
\end{subfigure}
\begin{subfigure}{0.09\textwidth}
\centering
5
\end{subfigure}
\begin{subfigure}{0.09\textwidth}
\centering
6
\end{subfigure}
\begin{subfigure}{0.09\textwidth}
\centering
7
\end{subfigure}
\begin{subfigure}{0.09\textwidth}
\centering
8
\end{subfigure}
\begin{subfigure}{0.09\textwidth}
\centering
9
\end{subfigure}
\caption{\textbf{The MNIST Dataset}}
\label{fig:mnist_img}
\end{figure*}

\begin{figure*}[!tb]
\centering
\begin{subfigure}{0.09\textwidth}
\centering
\includegraphics[width=0.9\linewidth]{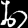}
\end{subfigure}
\begin{subfigure}{0.09\textwidth}
\centering
\includegraphics[width=0.9\linewidth]{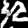}
\end{subfigure}
\begin{subfigure}{0.09\textwidth}
\centering
\includegraphics[width=0.9\linewidth]{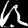}
\end{subfigure}
\begin{subfigure}{0.09\textwidth}
\centering
\includegraphics[width=0.9\linewidth]{figs/imgs/KMNIST_train_3.png}
\end{subfigure}
\begin{subfigure}{0.09\textwidth}
\centering
\includegraphics[width=0.9\linewidth]{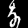}
\end{subfigure}
\begin{subfigure}{0.09\textwidth}
\centering
\includegraphics[width=0.9\linewidth]{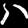}
\end{subfigure}
\begin{subfigure}{0.09\textwidth}
\centering
\includegraphics[width=0.9\linewidth]{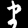}
\end{subfigure}
\begin{subfigure}{0.09\textwidth}
\centering
\includegraphics[width=0.9\linewidth]{figs/imgs/KMNIST_train_7.png}
\end{subfigure}
\begin{subfigure}{0.09\textwidth}
\centering
\includegraphics[width=0.9\linewidth]{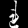}
\end{subfigure}
\begin{subfigure}{0.09\textwidth}
\centering
\includegraphics[width=0.9\linewidth]{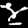}
\end{subfigure}

\par\bigskip
\begin{subfigure}{0.09\textwidth}
\centering
\includegraphics[width=0.9\linewidth]{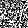}
\end{subfigure}
\begin{subfigure}{0.09\textwidth}
\centering
\includegraphics[width=0.9\linewidth]{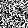}
\end{subfigure}
\begin{subfigure}{0.09\textwidth}
\centering
\includegraphics[width=0.9\linewidth]{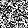}
\end{subfigure}
\begin{subfigure}{0.09\textwidth}
\centering
\includegraphics[width=0.9\linewidth]{figs/imgs/KMNIST_syn_3.png}
\end{subfigure}
\begin{subfigure}{0.09\textwidth}
\centering
\includegraphics[width=0.9\linewidth]{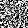}
\end{subfigure}
\begin{subfigure}{0.09\textwidth}
\centering
\includegraphics[width=0.9\linewidth]{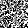}
\end{subfigure}
\begin{subfigure}{0.09\textwidth}
\centering
\includegraphics[width=0.9\linewidth]{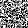}
\end{subfigure}
\begin{subfigure}{0.09\textwidth}
\centering
\includegraphics[width=0.9\linewidth]{figs/imgs/KMNIST_syn_7.png}
\end{subfigure}
\begin{subfigure}{0.09\textwidth}
\centering
\includegraphics[width=0.9\linewidth]{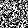}
\end{subfigure}
\begin{subfigure}{0.09\textwidth}
\centering
\includegraphics[width=0.9\linewidth]{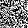}
\end{subfigure}

\par\bigskip
\begin{subfigure}{0.09\textwidth}
\centering
\begin{CJK}{UTF8}{min}
お
\end{CJK}
\end{subfigure}
\begin{subfigure}{0.09\textwidth}
\centering
\begin{CJK}{UTF8}{min}
き
\end{CJK}
\end{subfigure}
\begin{subfigure}{0.09\textwidth}
\centering
\begin{CJK}{UTF8}{min}
す
\end{CJK}
\end{subfigure}
\begin{subfigure}{0.09\textwidth}
\centering
\begin{CJK}{UTF8}{min}
つ
\end{CJK}
\end{subfigure}
\begin{subfigure}{0.09\textwidth}
\centering
\begin{CJK}{UTF8}{min}
な
\end{CJK}
\end{subfigure}
\begin{subfigure}{0.09\textwidth}
\centering
\begin{CJK}{UTF8}{min}
は
\end{CJK}
\end{subfigure}
\begin{subfigure}{0.09\textwidth}
\centering
\begin{CJK}{UTF8}{min}
ま
\end{CJK}
\end{subfigure}
\begin{subfigure}{0.09\textwidth}
\centering
\begin{CJK}{UTF8}{min}
や
\end{CJK}
\end{subfigure}
\begin{subfigure}{0.09\textwidth}
\centering
\begin{CJK}{UTF8}{min}
れ
\end{CJK}
\end{subfigure}
\begin{subfigure}{0.09\textwidth}
\centering
\begin{CJK}{UTF8}{min}
を
\end{CJK}
\end{subfigure}
\caption{\textbf{The KMNIST Dataset}}
\label{fig:kmnist_img}

\end{figure*}

We show the synthetic images using \textit{OPT-SYN} with the best substitute model. We compare them with the victim's training data in the SVHN, CIFAR-10, MNIST and KMNIST datasets (Figure~\ref{fig:svhn_img},~\ref{fig:cifar_img},~\ref{fig:mnist_img},~\ref{fig:kmnist_img}).
First row: images in the victim's training dataset. Second row: images in the synthetic dataset. Third Row: corresponding labels of images. 



